\documentclass[10pt,twocolumn,letterpaper]{article}

\usepackage{iccv}
\usepackage{times}
\usepackage{epsfig}
\usepackage{graphicx}
\usepackage{amsmath}
\usepackage{amssymb}
\usepackage{algorithm}
\usepackage{algorithmic}
\usepackage{booktabs}
\usepackage{subfigure}
\usepackage{amsfonts}
\usepackage{color}

\usepackage[breaklinks=true,bookmarks=false]{hyperref}

\iccvfinalcopy 

\def\iccvPaperID{****} 

\ificcvfinal\fi

\begin{document}

\title{Vector-Decomposed Disentanglement for Domain-Invariant Object Detection}

\author{Aming Wu$^{1,2}$\thanks{Equal contributions} \quad Rui Liu$^{1,2*}$ \quad Yahong Han$^{1,2,3}$\thanks{Corresponding author} \quad Linchao Zhu$^{4}$ \quad Yi Yang$^{4}$\\
$^{1}$College of Intelligence and Computing, Tianjin University, Tianjin, China\\
$^{2}$Tianjin Key Lab of Machine Learning, Tianjin University, Tianjin, China\\
$^{3}$Peng Cheng Laboratory, Shenzhen, China\\
$^{4}$ReLER Lab, AAII, University of Technology Sydney\\
\{tjwam, ruiliu, yahong\}@tju.edu.cn, \{Linchao.Zhu, yi.yang\}@uts.edu.au
}

\maketitle
\ificcvfinal\thispagestyle{empty}\fi

\begin{abstract}
   To improve the generalization of detectors, for domain adaptive object detection (DAOD), recent advances mainly explore aligning feature-level distributions between the source and single-target domain, which may neglect the impact of domain-specific information existing in the aligned features. Towards DAOD, it is important to extract domain-invariant object representations. To this end, in this paper, we try to disentangle domain-invariant representations from domain-specific representations. And we propose a novel disentangled method based on vector decomposition. Firstly, an extractor is devised to separate domain-invariant representations from the input, which are used for extracting object proposals. Secondly, domain-specific representations are introduced as the differences between the input and domain-invariant representations. Through the difference operation, the gap between the domain-specific and domain-invariant representations is enlarged, which promotes domain-invariant representations to contain more domain-irrelevant information. In the experiment, we separately evaluate our method on the single- and compound-target case. For the single-target case, experimental results of four domain-shift scenes show our method obtains a significant performance gain over baseline methods. Moreover, for the compound-target case (i.e., the target is a compound of two different domains without domain labels), our method outperforms baseline methods by around 4\%, which demonstrates the effectiveness of our method. 
\end{abstract}

\section{Introduction}

\begin{figure}
\centering
\includegraphics[width=1.0\linewidth]{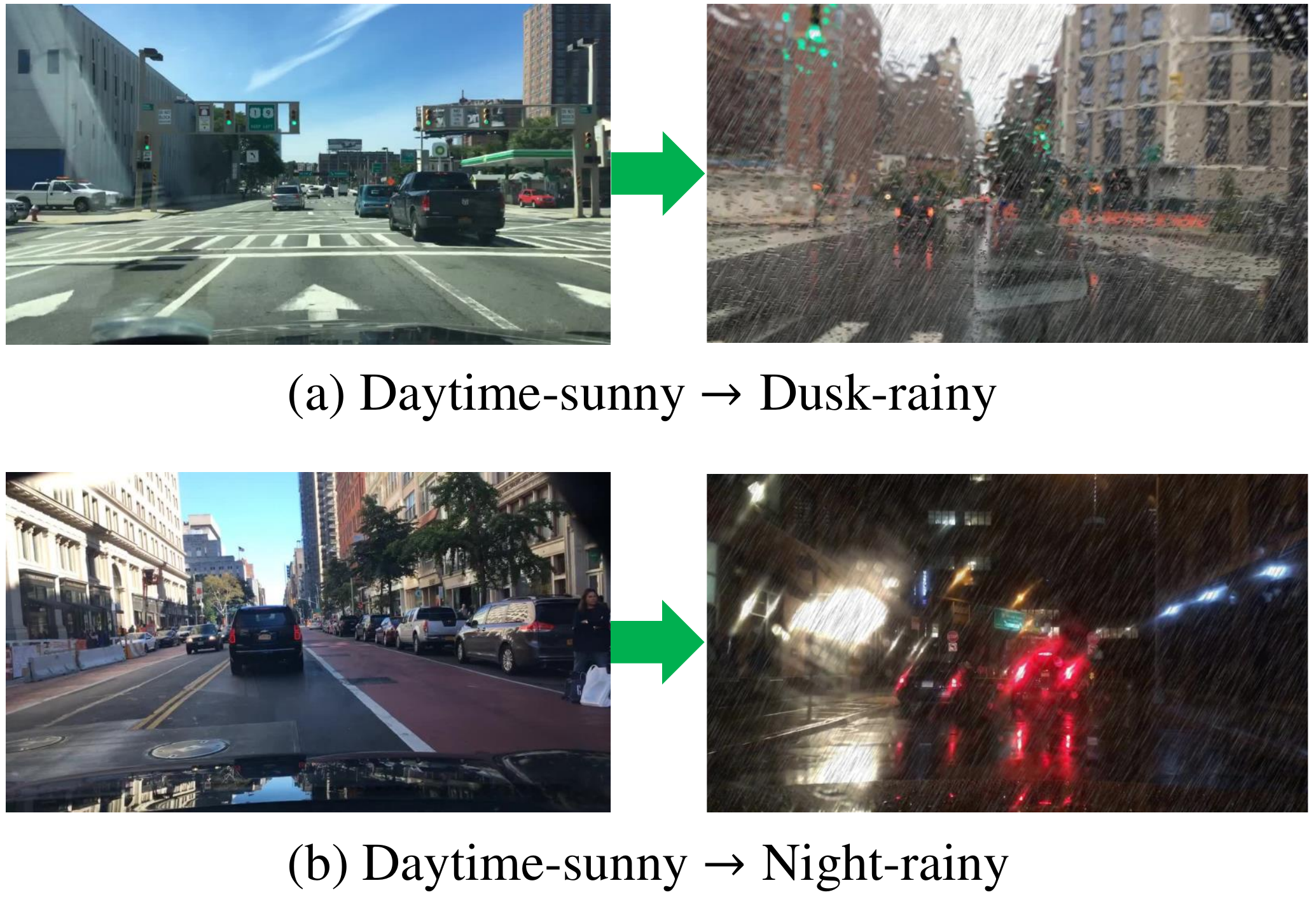}
\caption{To further verify the proposed method, we construct two new adaptive scenes with different weather conditions.}
\label{fig1}
\vspace{-0.15in}
\end{figure}

Though object detection has achieved many advances \cite{ren2015faster,he2017mask,MM3,MM5,redmon2016you,liu2016ssd}, when the training and test data are from different domains, these methods usually suffer from poor generalization. To this end, the task of domain adaptive object detection (DAOD) \cite{chen2018domain} has been proposed, in which a domain gap always exists between the training/source and test/target domain, e.g., different weather conditions (as shown in Fig. \ref{fig1}).

To address DAOD, many methods \cite{shao2018feature,Li,wang2018visual,zhuo2017deep} explored to reduce the domain gap by aligning the feature-level distribution of the source and single-target domain, which may neglect the impact of the domain-specific information existing in the aligned features. Towards DAOD, it is important to obtain domain-invariant representations (DIR), which is a bridge to alleviate the domain-shift impact and can help extract domain-invariant object features.

In this paper, we focus on extracting DIR. We explore to employ disentangled representation learning (DRL) \cite{bengio2013representation,locatello2018challenging} to disentangle DIR from domain-specific representations (DSR). As a method of feature decomposition, the purpose of DRL is to uncover a set of independent factors that give rise to the current observation \cite{do2019theory}. And these factors should contain all the information in the observation. Inspired by the idea, we explore to utilize DRL to solve DAOD and propose a novel disentangled method to extract DIR.
\begin{figure}
\centering
\includegraphics[width=0.95\linewidth]{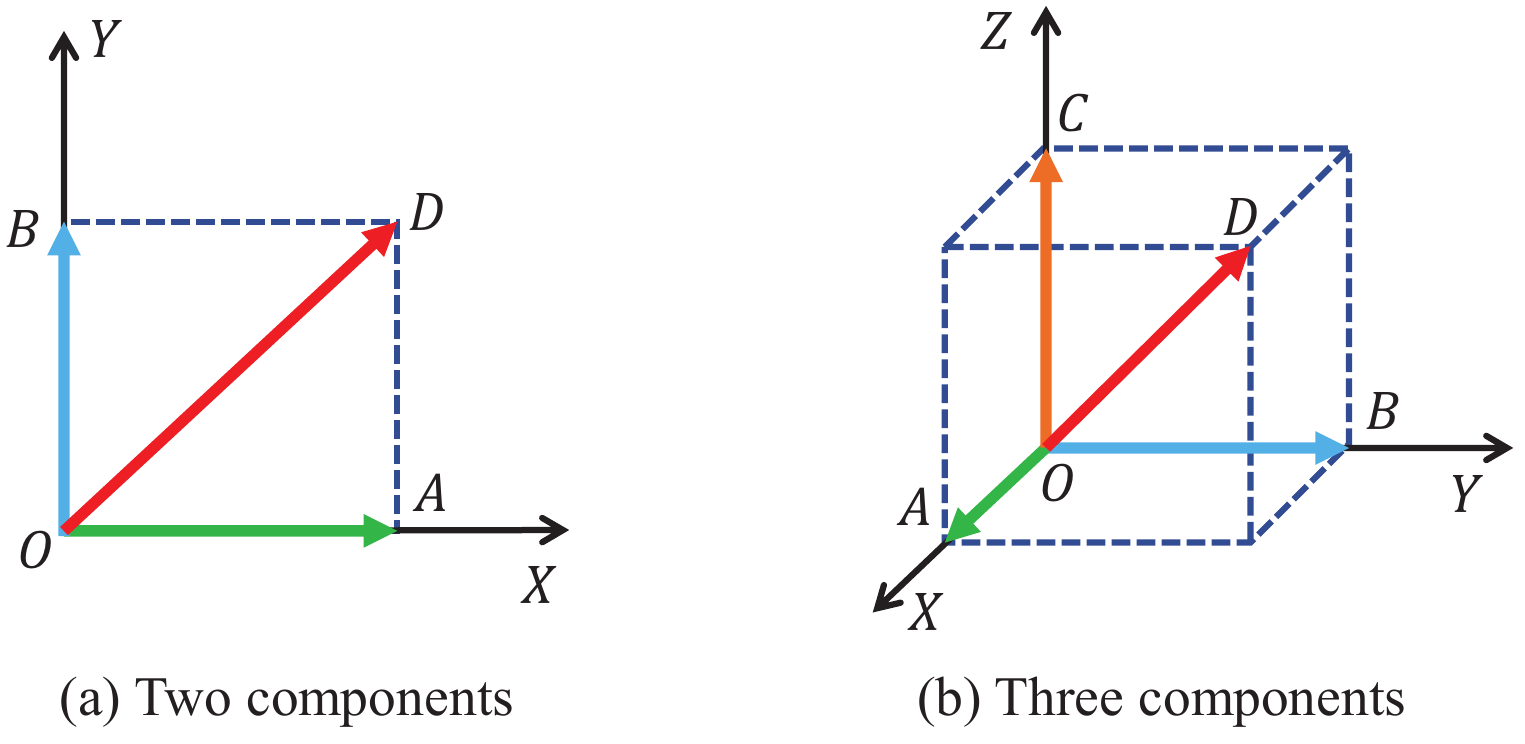}
\caption{Two examples of vector decomposition. (a) is the case of two components (i.e., $\protect\overrightarrow{OD}$ = $\protect\overrightarrow{OA}$ + $\protect\overrightarrow{OB}$). (b) is the case of three components (i.e., $\protect\overrightarrow{OD}$ = $\protect\overrightarrow{OA}$ + $\protect\overrightarrow{OB}$ + $\protect\overrightarrow{OC}$). Here, all these components are kept orthogonal.}
\label{fig2}
\vspace{-0.1in}
\end{figure}
Particularly, we cast DRL into a process of vector decomposition. Vector decomposition is the general process of breaking one vector into two or more vectors that add up to the original vector, which is similar in spirit to the process of disentanglement \cite{higgins2018towards}. Thus we consider employing the idea of vector decomposition to conduct disentanglement.

Concretely, given a feature map extracted by a backbone, an extractor consisting of multiple convolutional layers is devised to separate DIR from the feature map. Next, we take the difference between the feature map and DIR as DSR. Meanwhile, a domain classifier is used to help DSR contain much more domain-specific information. Besides, one key-step of disentanglement is to keep DIR and DSR independent. In this paper, we enhance independence via a constraint of vector orthogonalization between the DIR and DSR. Finally, a region proposal network (RPN) is utilized to extract object proposals from DIR. Moreover, since the proposed method is a new feature decomposition mechanism, we should design a proper optimization to obtain DIR. To this end, based on the purpose of DRL, we break DRL into two sequential training steps, i.e., the step of feature decomposition aiming at learning disentanglement, and the step of feature orthogonalization aiming at promoting DIR and DSR to be independent. The two-step optimization could promote our model learns feature decomposition, which is beneficial for extracting DIR for DAOD.

In the experiment, we first evaluate our method on the single-target case. Next, we evaluate our method on the compound-target case \cite{liu2020open}, i.e., the target is a compound of two different domains without domain labels. The significant performance gain over baselines shows the effectiveness of our disentangled method. Our code will be available at \url{https://github.com/AmingWu/VDD-DAOD}.

The contributions are summarized as follows:

(1) Different from traditional disentanglement, we present a vector-decomposed disentanglement, which does not rely on the reconstruction operation to ensure the decomposed components contain all the information of input.

(2) Based on vector-decomposed disentanglement, we design a new framework to solve DAOD. Meanwhile, we design a two-step training strategy to optimize our model.

\begin{figure}
\centering
\includegraphics[width=1.0\linewidth]{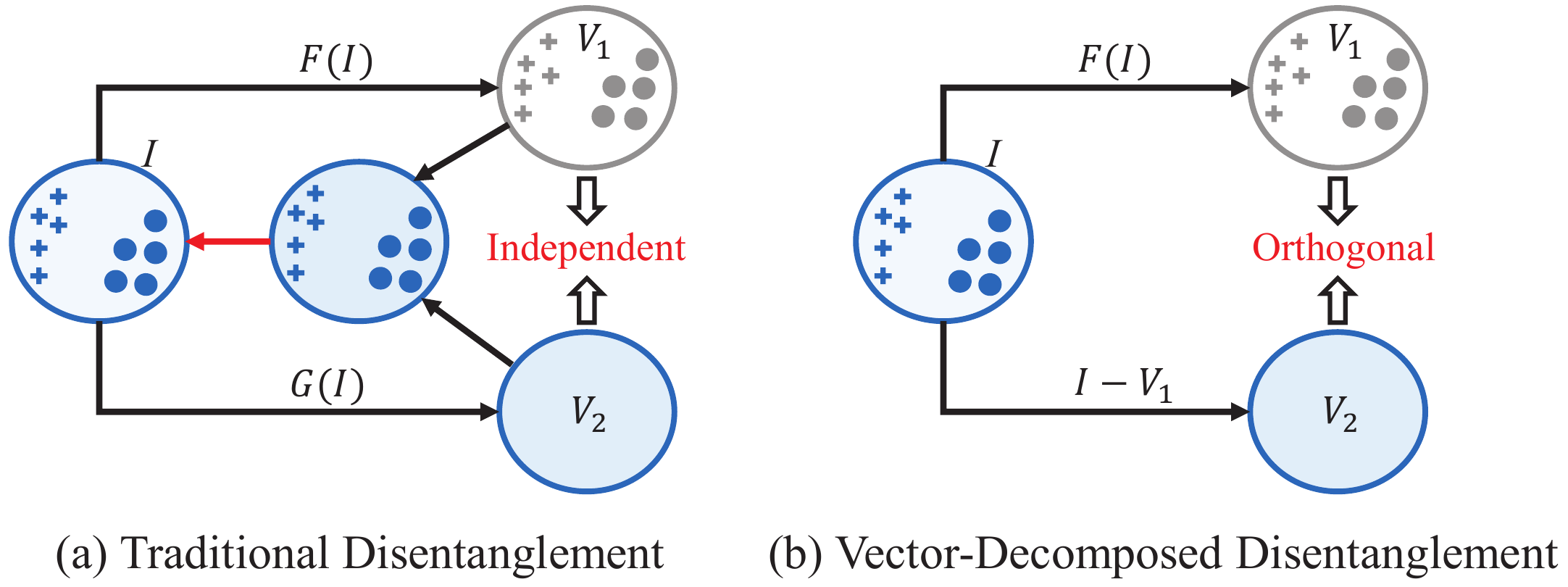}
\caption{Comparisons between the traditional method and our vector-decomposed method. Given an input $I$, traditional disentanglement usually employs two extractors $F$ and $G$ to disentangle $V_{1}$ and $V_{2}$. And $V_{1}$ and $V_{2}$ should be kept independent. To promote $V_{1}$ and $V_{2}$ to contain all the information of $I$, a reconstruction operation is usually employed. Here, the red arrow indicates the reconstruction operation. For vector-decomposed disentanglement, it only uses an extractor to decompose $V_{1}$. The difference between $I$ and $V_{1}$ is taken as $V_{2}$. Meanwhile, $V_{1}$ and $V_{2}$ are kept orthogonal. Besides, vector-decomposed disentanglement does not need to utilize the reconstruction operation to promote $V_{1}$ and $V_{2}$ to contain all the information of $I$.}
\label{fig3}
\vspace{-0.1in}
\end{figure}

(3) In the experiment, our method is separately evaluated on the single- and compound-target cases. And we build two new adaptive scenes (see Fig. \ref{fig1}), i.e., Daytime-sunny $\rightarrow$ Dusk-rainy and Daytime-sunny $\rightarrow$ Night-rainy, to further verify our method. The significant performance gain over baselines shows the effectiveness of our method.

\section{Related Work}

\textbf{Domain Adaptive Object Detection.} Most existing methods \cite{xu2020exploring,chen2020harmonizing,xu2020cross,shen2019scl,zhao2020adaptive,zhao2020collaborative} employ holistic representations to align the feature- or pixel-level distributions of the source and target domain. Particularly, Chen et al. \cite{chen2018domain} proposed to align the global feature distributions to reduce the domain gap. Saito et al. \cite{saito2019strong} proposed to align the local and global feature distributions to alleviate the domain-shift impact. Besides, the work \cite{kim2019diversify} utilized an encoder-decoder network to translate the style of the source domain to that of the target domain, which could be thought of as aligning the pixel-level distributions of the source and target domain. Although these methods have been demonstrated to be effective, they neglect the impact of domain-specific information existing in the aligned features, which may affect the adaptation performance. To this end, we focus on extracting domain-invariant representations for DAOD.

\begin{figure*}
\centering
\includegraphics[width=0.9\linewidth]{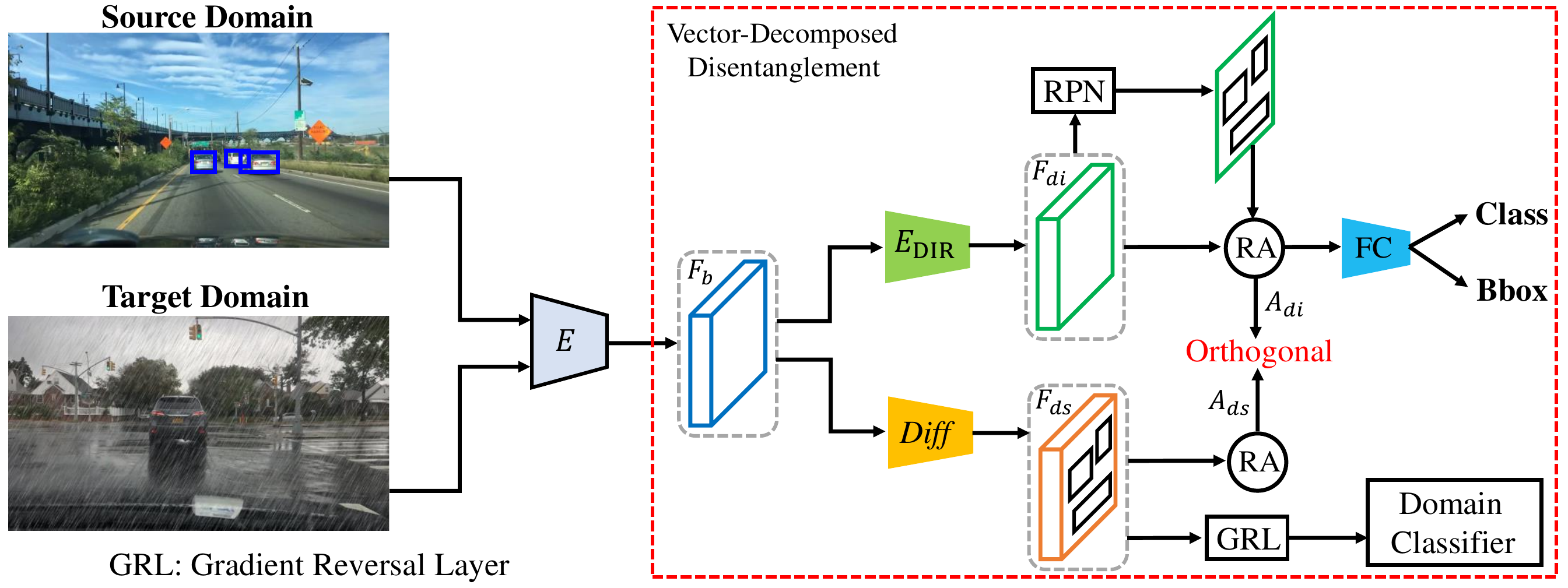}
\caption{Illustration of vector-decomposed disentanglement: a plug component for domain adaptive Faster R-CNN series \cite{saito2019strong,xu2020exploring}. `RA' and `$Diff$' separately indicate Roi-Alignment and the difference decomposition. `GRL' denotes Gradient Reversal Layer \cite{chen2018domain,saito2019strong}. We first design an extractor $E_{\rm DIR}$ to decompose DIR ($F_{di}$) from $F_{b}$. Then, based on $F_{di}$, RPN is employed to extract domain-invariant proposals.}
\label{fig4}
\vspace{-0.1in}
\end{figure*}

\textbf{Disentangled Representation Learning.} As an effective mechanism of feature decomposition, recently, DRL \cite{liu2018detach,ijcai2019-285} has been demonstrated to be effective in many tasks, e.g., image translation \cite{lee2018diverse} and few-shot learning \cite{ridgeway2018learning}. Particularly, the work \cite{lee2018diverse} employs DRL to decompose DSR to make diverse image style translation. Peng et al. \cite{peng2019domain} utilize DRL to disentangle three different factors to make domain adaptive classification. However, since this work only considers holistic image-level representations for classification, it could not be applied directly to object detection.

In this paper, we consider DRL from the perspective of vector decomposition. Particularly, our method only requires to devise an extractor to decompose DIR. And DSR could be obtained from the difference between the input and DIR. Experimental results on single- and compound-target DAOD demonstrate the effectiveness of our method.

\section{Vector-Decomposed Disentanglement}

As discussed in the section of Introduction, the purpose of vector decomposition is to break one vector into two or more components that add up to the original vector. In general, each vector can be taken as the sum of two or more other vectors. Fig. \ref{fig2} shows two decomposed examples, i.e., $\protect\overrightarrow{OD} = \protect\overrightarrow{OA} + \protect\overrightarrow{OB}$ and $\protect\overrightarrow{OD} = \protect\overrightarrow{OA} + \protect\overrightarrow{OB} + \protect\overrightarrow{OC}$.

Obviously, vector decomposition is similar in spirit to disentanglement. And the decomposition idea is also applied to high-dimensional space. Therefore, we consider employing vector decomposition to obtain disentangled representations. Concretely, for the case of two components (Fig. \ref{fig3}(b)), give an input representation $I$, we design an extractor $F$ to decompose the first component $V_{1}$ from $I$. Then, we take the difference between $I$ and $V_{1}$ as the second component $V_{2}$. Here, we name the process extracting $V_{2}$ as difference decomposition.
\begin{equation}\label{eq1}
V_{1} = F(I), \; V_{2} = I - V_{1}, \; V_{1} \bot V_{2},
\end{equation}
\noindent where $\bot$ indicates two components are orthogonal.

Compared with the traditional disentanglement (Fig. \ref{fig3}(a)), vector decomposition only takes the difference between the original input and decomposed components as the last component, which reduces parameters and computational costs. Moreover, the difference decomposition of obtaining the last component could make all the components contain all the information of the input, which does not rely on the reconstruction operation. In the following, we will introduce the details of vector-decomposed disentanglement for domain adaptive object detection.

\section{Domain-Invariant Object Detection}

For DAOD, we could access image $x^{s}$ with labels $y^{s}$ and bounding boxes $b^{s}$, which are from the source domain. And we could also access image $x^{t}$ that is from the target domain. The goal is to obtain the results of the target domain.

\subsection{The Network of Disentanglement}

The right part of Fig. \ref{fig4} illustrates the details of vector-decomposed disentanglement, which is plugged into the domain adaptive Faster R-CNN series \cite{saito2019strong,xu2020exploring,ren2015faster}. Concretely, given an image $x^{s}$ and $x^{t}$, we first obtain a feature map $F_{b}$ that is the output of a feature extractor $E$. Next, we define an extractor $E_{\rm DIR}$ to decompose domain-invariant feature $F_{di}$ from $F_{b}$. And the difference between $F_{b}$ and $F_{di}$ is taken as the domain-specific feature $F_{ds}$.
\begin{equation}\label{eq3}
F_{di} = E_{\rm DIR}(F_{b}),\; F_{ds} = F_{b} - F_{di}.
\end{equation}

Here, $E_{\rm DIR}$ indicates the DIR extractor. The size of $F_{di}$ and $F_{ds}$ is set to the same as that of $F_{b}$. Next, a Region Proposal Network (RPN) is performed on $F_{di}$ to extract a set of domain-invariant proposals. Finally, for an image from the source domain, the detection loss is defined as follows:
\begin{equation}\label{eq4}
\mathcal{L}_{det} = \mathcal{L}_{loc} + \mathcal{L}_{cls} + \mathcal{L}_{rpn},
\end{equation}
\noindent where $\mathcal{L}_{loc}$ and $\mathcal{L}_{cls}$ separately indicate the bounding-box regression loss and classification loss. $\mathcal{L}_{rpn}$ is the loss of RPN to distinguish foreground from background and to refine bounding-box anchors.

\begin{figure*}
\centering
\includegraphics[width=1.0\linewidth]{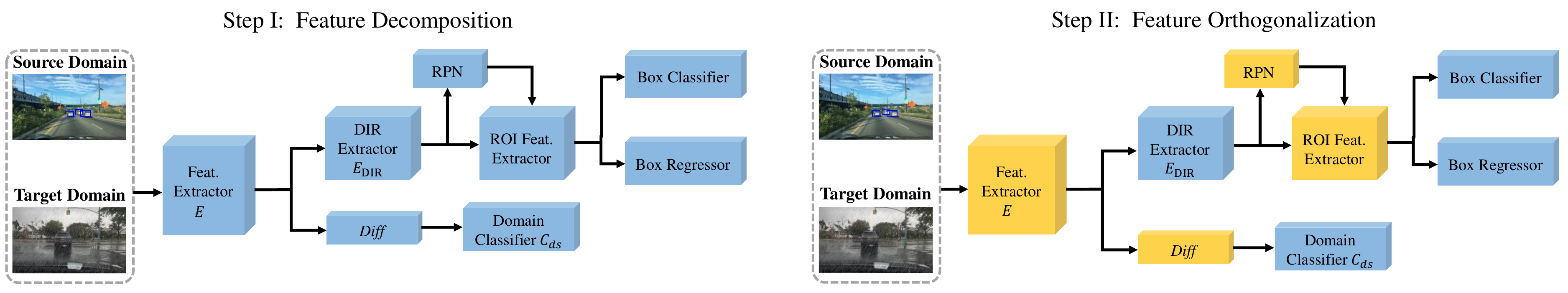}
\caption{Illustration of our two-step optimization process. In the first training step (i.e., feature decomposition), the entire object detector (all the blue blocks) are jointly trained on the source and target images. In the second training step (i.e, feature orthogonalization), the parameters in the yellow blocks are fixed. We only update the parameters in the blue blocks.}
\label{fig5}
\vspace{-0.1in}
\end{figure*}

\subsection{Training with the Two-step Optimization}

The goal of our method (see Eq. \eqref{eq1}) is to decompose a set of orthogonal components. To enhance the disentangled ability, we break vector decomposition into two sequential steps. Specifically, we first promote models to be capable of decomposing components. Then, a constraint is imposed to promote these components to be orthogonal.

\textbf{The step of feature decomposition.} The step is to promote our model to decompose input features into two different components. Concretely, based on $F_{di}$, we first employ RPN to extract object proposals. Then, for a source image, the processes of detection loss are shown in Eq. \eqref{eq4}.

Next, to promote the difference result $F_{ds}$ to contain much more domain-specific information, we utilize the adversarial training mechanism \cite{ganin2014unsupervised} and design a network $C_{ds}$ to perform domain classification. And the domain label $D$ is set to 0 for the source domain and 1 for the target domain. Finally, the loss of the first step is shown as follows:
\begin{equation}\label{eq5}
\begin{aligned}
\mathcal{L}_{src}^{1}&=\mathcal{L}_{det} + \mathcal{L}_{dom}(C_{ds}(F_{ds})), \\
\mathcal{L}_{tgt}^{1}&=\mathcal{L}_{dom}(C_{ds}(F_{ds})),
\end{aligned}
\end{equation}
\noindent where $\mathcal{L}_{src}^{1}$ and $\mathcal{L}_{tgt}^{1}$ are the objective functions of the source and target domain, respectively. $\mathcal{L}_{dom}$ is the domain classification loss, i.e., $\mathcal{L}_{dom} = -[D{\rm log}\hat{D}+(1-D){\rm log}(1-\hat{D})]$ and $\hat{D} = C_{ds}(F_{ds})$. Finally, we take the sum of $\mathcal{L}_{src}^{1}$ and $\mathcal{L}_{tgt}^{1}$ to optimize the entire model.

\textbf{The step of feature orthogonalization.} In this step, we first fix the feature extractor $E$. Then, we use the extractor $E_{\rm DIR}$ to obtain $F_{di}$ (Eq. \eqref{eq3}). Next, RPN is performed on $F_{di}$ to extract a set of object proposals.

The key idea of disentanglement \cite{do2019theory} is to keep the disentangled components independent. Here, based on the theory of vector decomposition, we try to promote the decomposed components are orthogonal, which is equivalent to the independent operation. Thus we impose an \textbf{orthogonal loss} $\mathcal{L}_{\perp}$ on the DIR and DSR. Concretely, based on object proposals, we first obtain the Roi-Alignment result $A_{di} \in \mathbb{R}^{n \times c \times h \times w}$ of $F_{di}$ and $A_{ds} \in \mathbb{R}^{n \times c \times h \times w}$ of $F_{ds}$, where $n$, $c$, $h$, and $w$ indicate the number of proposals, the number of channels, the height and width, respectively. The process of orthogonal loss is shown as follows:
\begin{equation}\label{eq6}
\begin{aligned}
& M = (||P_{di}||_{2}^{2}) \odot (||P_{ds}||_{2}^{2}), \\
& \mathcal{L}_{\perp} = \frac{1}{n}\sum_{i=1}^{n}\mid\sum_{j=1}^{c}M[i,j]\mid,
\end{aligned}
\end{equation}
\noindent where $P_{di} \in \mathbb{R}^{n \times c}$ and $P_{ds} \in \mathbb{R}^{n \times c}$ are the results of global average pooling. $|| \cdot ||_{2}^{2}$, $\mid \cdot \mid$, and $\odot$ separately indicate L2-norm, the absolute value operation, and element-wise product. $M[i,j]$ indicates the value of $M \in \mathbb{R}^{n \times c}$ at the position $(i,j)$. Besides, it is worth noting that we use the alignment results instead of the overall feature map to compute the orthogonal loss, which could not only reduce computational costs but also promote our model to focus on object regions.

By minimizing the orthogonal loss, we could promote $F_{di}$ and $F_{ds}$ are independent. Since $F_{ds}$ contains more domain-specific information, this loss can promote $F_{di}$ to contain much more domain-invariant information. Finally, the loss of the second step is defined as follows:
\begin{equation}\label{eq7}
\begin{aligned}
& \mathcal{L}_{src}^{2} = \mathcal{L}_{det} + \mathcal{L}_{dom}(C_{ds}(F_{ds})) + \mathcal{L}_{\perp}, \\
& \mathcal{L}_{tgt}^{2} = \mathcal{L}_{dom}(C_{ds}(F_{ds})) + \mathcal{L}_{\perp},
\end{aligned}
\end{equation}
\noindent where $\mathcal{L}_{det}$ is the detection loss based on $A_{di}$. The sum of $\mathcal{L}_{src}^{2}$ and $\mathcal{L}_{tgt}^{2}$ is used to optimize certain components of the model. The processes are shown in the right part of Fig. \ref{fig5}. After the second training step, the decomposed DIR and DSR will be kept independent, which enhances the disentangled ability of our model.

In this paper, our model is trained in an end-to-end way. The training details are shown in Algorithm \ref{alg_DAOD}. Besides, for the second training step, the parameters that do not appear in the step are considered to be fixed.

\subsection{Discussion about Learning DIR}

For our method, we have two operations to promote to learn domain-invariant features. Firstly, the difference decomposition makes $F_{di}$ contain much less domain-relevant information. Secondly, the orthogonal loss can further promote $F_{di}$ to contain much more domain-irrelevant information. And we consider domain-irrelevant information contains domain-invariant information. Thus, these two operations promote $F_{di}$ contains much more domain-invariant information, which reduces the domain-shift impact.

\begin{algorithm}[t]
  \caption{Two-step optimization for DAOD} \label{alg_DAOD}
  \begin{small}
      \begin{algorithmic}[1]
          \REQUIRE~~ \\
          source images $\{x^{s}, y^{s}, b^{s}\}$; target images $\{x^{t}\}$; feature extractor ${E}$; DIR extractor ${E}_{\rm DIR}$; domain classifier ${C}_{ds}$. \\
          \ENSURE~~ \\
          feature extractor $\hat{E}$, DIR extractor ${\hat{E}}_{\rm DIR}$. \\
          \WHILE {not converged}
          \STATE  Sample a mini-batch from $\{x^{s}, y^{s}, b^{s}\}$ and $\{x^{t}\}$;
          \STATE \textbf{Feature Decomposition:}
          \STATE Compute $\mathcal{L}_{1} = \mathcal{L}_{src}^{1} + \mathcal{L}_{tgt}^{1}$ (Eq. \eqref{eq5});
          \STATE Update ${E}$, ${E}_{\rm DIR}$, and ${C}_{ds}$ by $\mathcal{L}_{1}$;
          \STATE Update RPN module, Classifier, and Regressor by $\mathcal{L}_{1}$;
          \STATE \textbf{Feature Orthogonalization:}
          \STATE Compute $\mathcal{L}_{2} = \mathcal{L}_{src}^{2} + \mathcal{L}_{tgt}^{2}$ (Eq. \eqref{eq7});
          \STATE Update ${E}_{\rm DIR}$, ${C}_{ds}$ by $\mathcal{L}_{2}$;
          \STATE Update Classifier and Regressor by $\mathcal{L}_{2}$;
          \ENDWHILE
          \RETURN $\hat{E}={E}; \hat{E}_{\rm DIR}={E}_{\rm DIR}$.
      \end{algorithmic}
    \end{small}
\end{algorithm}

\section{Experiment}

In the experiment, we separately evaluate our approach on single- and compound-target DAOD. For the single-target case, our method is evaluated on four domain-shift scenes, i.e., Cityscapes \cite{cordts2016cityscapes} $\rightarrow$ FoggyCityscapes \cite{sakaridis2018semantic}, PASCAL \cite{everingham2010pascal} $\rightarrow$ Watercolor \cite{inoue2018cross}, Daytime-sunny $\rightarrow$ Dusk-rainy, and Daytime-sunny $\rightarrow$ Night-rainy. For the compound-target case \cite{liu2020open}, we take Daytime-sunny as the source domain and the compound of Dusk-rainy and Night-rainy as the target domain, whose goal is to adapt a model from labeled source domain to unlabeled compound target domain. All the experiments are trained in an end-to-end way.

\textbf{Datasets.} Cityscapes is a dataset about city street scene. It contains 2,975 images for training and 500 images for validation. FoggyCityscapes is rendered based on Cityscapes. And it shows street scene under foggy weather. We follow the setting of the work \cite{saito2019strong} and evaluate our method on the validation set. For PASCAL $\rightarrow$ Watercolor, we utilize Pascal VOC dataset as the source domain. It contains 20 classes of images and bounding box annotations. Following the setting of the work \cite{saito2019strong}, we employ Pascal VOC 2007 and 2012 training and validation splits for training, which results in about 15K images. Watercolor contains 2K images with 6 categories. The splits of the training and test set are the same as the work \cite{saito2019strong}.

\begin{table}
\small
\begin{center}
\scalebox{0.80}{
\tabcolsep=3.3pt
\begin{tabular}{l|cccccccc|c}
\toprule[1.5pt]
Method  & prsn & rider & car & truck & bus & train & mcycl & bcycl & mAP \\
\hline
Source Only & 24.7 & 31.9 & 33.1 & 11.0 & 26.4 & 9.2 & 18.0 & 27.9 & 22.8 \\
\hline
DAF \cite{chen2018domain} & 25.0 & 31.0 & 40.5 & 22.1 & 35.3 & 20.2 & 20.0 & 27.1 & 27.6 \\ \hline
DT \cite{inoue2018cross} & 25.4 & 39.3 & 42.4 & 24.9 & 40.4 & 23.1 & 25.9 & 30.4 & 31.5 \\ \hline
SC-DA \cite{zhu2019adapting} & 33.5 & 38.0 & 48.5 & 26.5 & 39.0 & 23.3 & 28.0 & 33.6 & 33.8 \\ \hline
DMRL \cite{kim2019diversify} & 30.8 & 40.5 & 44.3 & 27.2 & 38.4 & 34.5 & 28.4 & 32.2 & 34.6 \\ \hline
MLDA \cite{xie2019multi} & 33.2 & 44.2 & 44.8 & 28.2 & 41.8 & 28.7 & 30.5 & 36.5 & 36.0 \\ \hline
FSDA \cite{wang2019few} & 29.1 & 39.7 & 42.9 & 20.8 & 37.4 & 24.1 & 26.5 & 29.9 & 31.3 \\ \hline
MAF \cite{he2019multi} & 28.2 & 39.5 & 43.9 & 23.8 & 39.9 & 33.3 & 29.2 & 33.9 & 34.0 \\ \hline
CT \cite{zhao2020collaborative} & 32.7 & 44.4 & 50.1 & 21.7 & 45.6 & 25.4 & 30.1 & 36.8 & 35.9 \\ \hline
CDN \cite{su2020adapting} & {\bf 35.8} & 45.7 & 50.9 & 30.1 & 42.5 & 29.8 & 30.8 & 36.5 & 36.6 \\ \hline
SCL \cite{shen2019scl} & 31.6 & 44.0 & 44.8 & 30.4 & 41.8 & 40.7 & 33.6 & 36.2 & 37.9 \\ \hline
ATF \cite{he2020domain} & 34.6 & 47.0 & 50.0 & 23.7 & 43.3 & 38.7 & 33.4 & {\bf 38.8} & 38.7 \\ \hline
MCAR \cite{zhao2020adaptive} & 32.0 & 42.1 & 43.9 & 31.3 & 44.1 & 43.4 & {\bf 37.4} & 36.6 & 38.8 \\ \hline
HTCN \cite{chen2020harmonizing} & 33.2 & {\bf 47.5} & 47.9 & 31.6 & 47.4 & {\bf 40.9} & 32.3 & 37.1 & 39.8 \\ \hline \hline
SW \cite{saito2019strong} & 29.9 & 42.3 & 43.5 & 24.5 & 36.2 & 32.6 & 30.0 & 35.3 & 34.3 \\
SW-VDD (ours) & 32.1 & 42.8 & 49.4 & 29.0 & 49.0 & 33.9 & 29.9 & 37.1 & 37.9 \\
ICCR \cite{xu2020exploring} & 32.9 & 43.8 & 49.2 & 27.2 & 45.1 & 36.4 & 30.3 & 34.6 & 37.4 \\
ICCR-VDD (ours) & 33.4 & 44.0 & {\bf 51.7} & {\bf 33.9} & {\bf 52.0} & 34.7 & 34.2 & 36.8 & {\bf 40.0} \\
\bottomrule[1.5pt]
\end{tabular}}
\end{center}
\vspace{-0.1in}
\caption{Results (\%) on adaptation from Cityscapes to FoggyCityscapes. `prsn', `mcycl', and `bcycl' separately denote `person', `motorcycle', and `bicycle' category. `VDD' indicates vector-decomposed disentanglement.}\label{table1}
\vspace{-0.15in}
\end{table}

The Berkeley Deep Drive 100k (BDD-100k) dataset \cite{yu2018bdd100k} consists of 100,000 driving videos. Based on this dataset, we build two new adaptive scenes. As shown in Fig. \ref{fig1}, for Daytime-sunny $\rightarrow$ Dusk-rainy, we select 27,708 daytime-sunny images as the source domain and 3,501 dusk-rainy images as the target domain. For Daytime-sunny $\rightarrow$ Night-rainy, we select 27,708 daytime-sunny images as the source domain and 2,494 night-rainy images as the target domain. Besides, for the compound-target case, we select 27,708 daytime-sunny images as the source domain and 5,995 images consisting of dusk-rainy and night-rainy as the compound target domain. Meanwhile, we render these rainy images to enlarge the gap between the source and target domain. The number of annotation boxes is around 455,000. We evaluate the performance on the target domain. Besides, the BDD-100k dataset includes ten categories. Here, we choose seven commonly used categories, which do not include the category of light, sign, and train.

\textbf{Implementation Details.} We employ three convolutional layers as the domain-invariant feature extractor $E_{\rm DIR}$. And we separately design a network with three fully connected layers as the domain classifiers. Finally, during training, we first train our model with learning rate 0.001 for 50K iterations, then with the learning rate 0.0001 for 30K more iterations. In the test, we utilize mean average precisions (mAP) as the evaluation metric. More details can be seen in the supplementary material.

\begin{figure*}[ht]
\begin{center}
  \subfigure[\footnotesize Raw image]{
  \begin{minipage}[t]{0.18\linewidth}
    \includegraphics[width=1.3in]{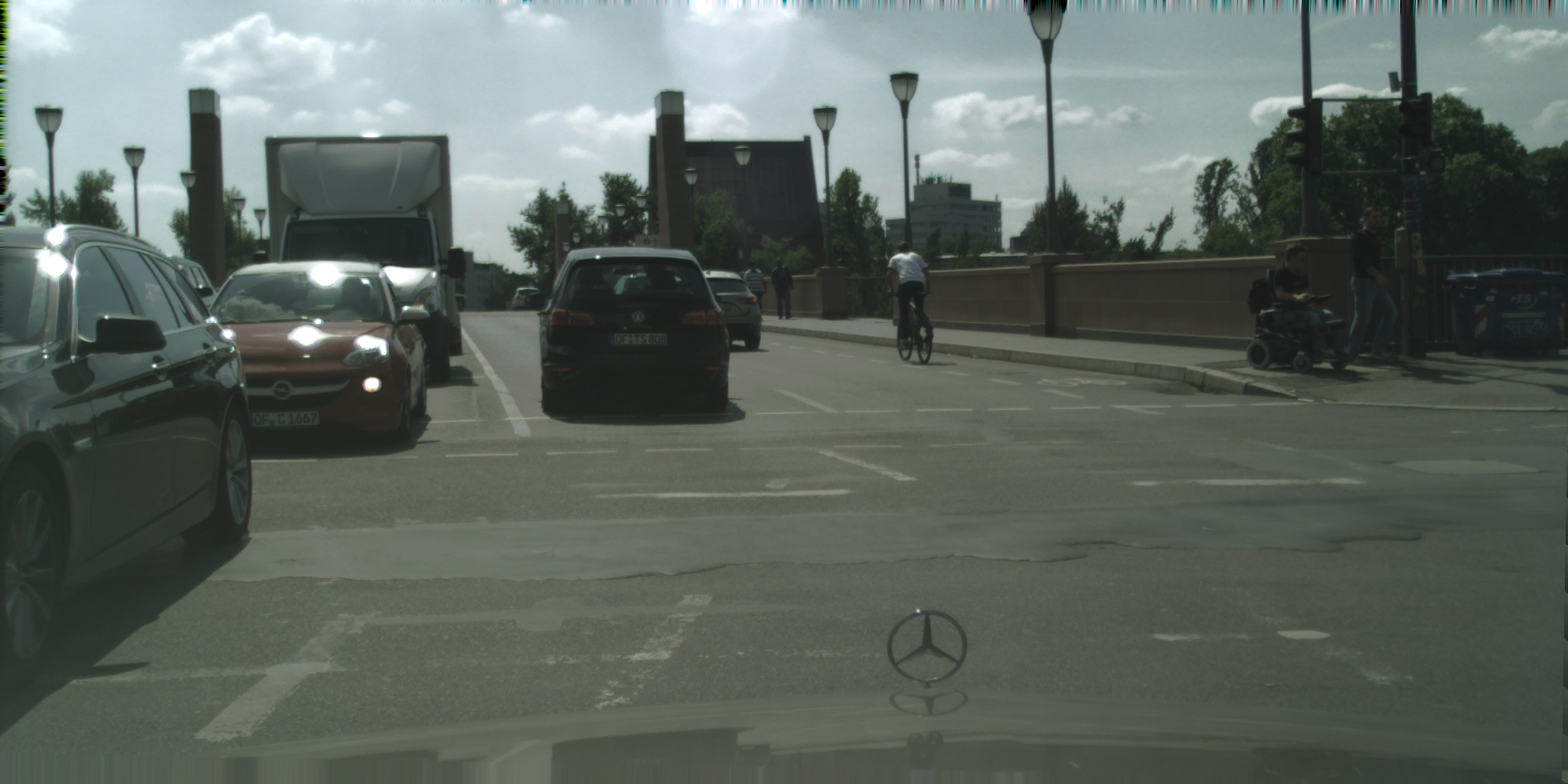}\\ \vspace{-0.1in}
    \includegraphics[width=1.3in]{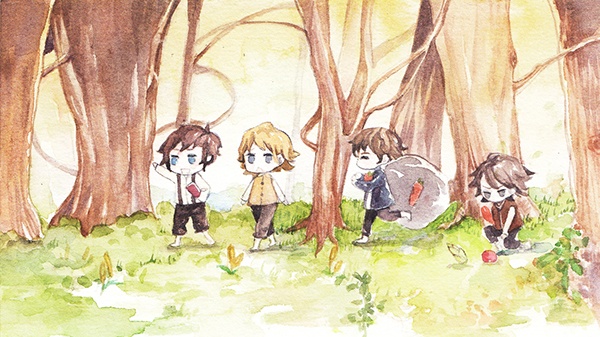}
  \end{minipage}
  }
  \hspace{0.01in}
  \subfigure[\footnotesize GT]{
  \begin{minipage}[t]{0.18\linewidth}
    \includegraphics[width=1.3in]{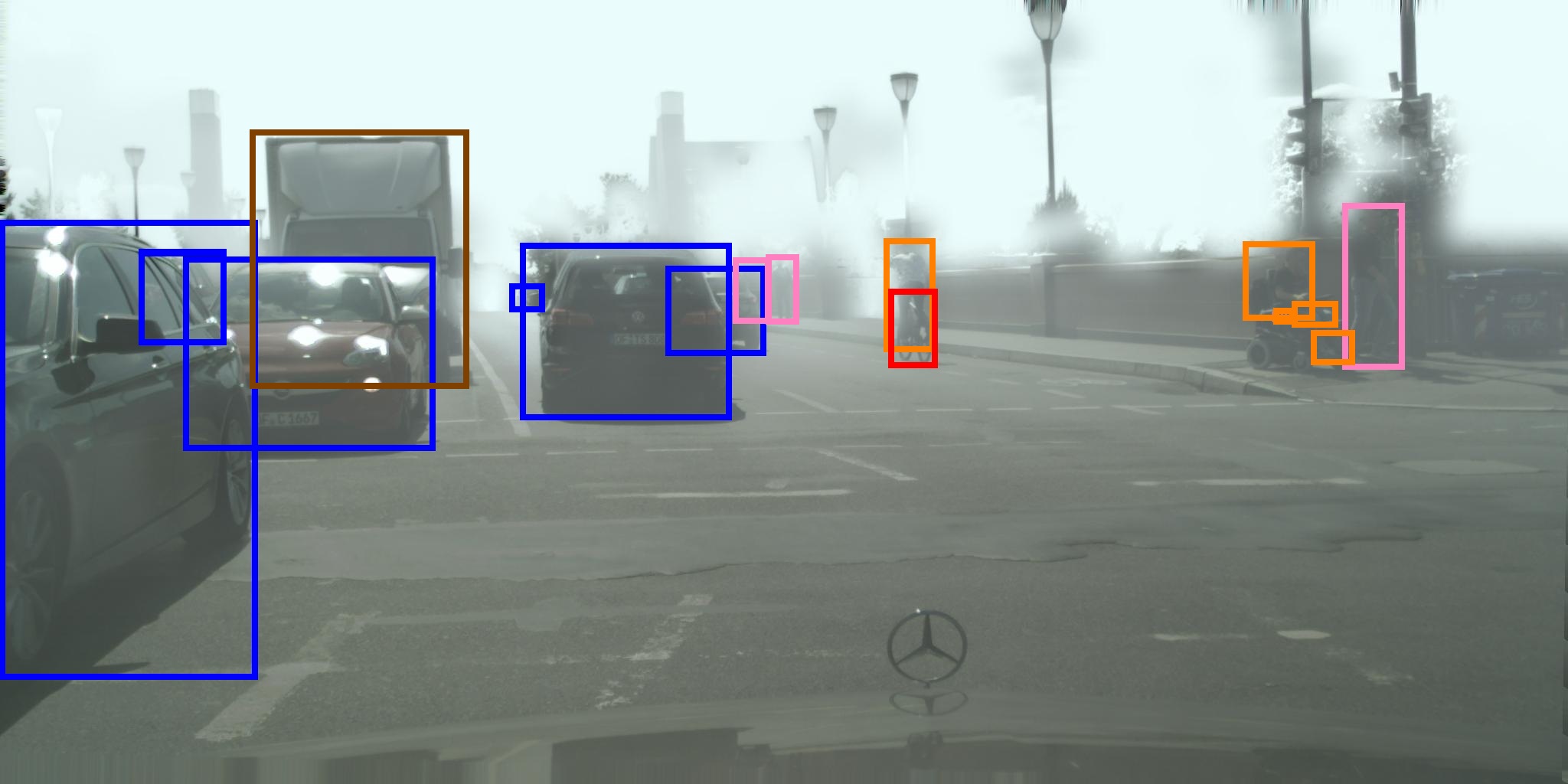}\\ \vspace{-0.1in}
    \includegraphics[width=1.3in]{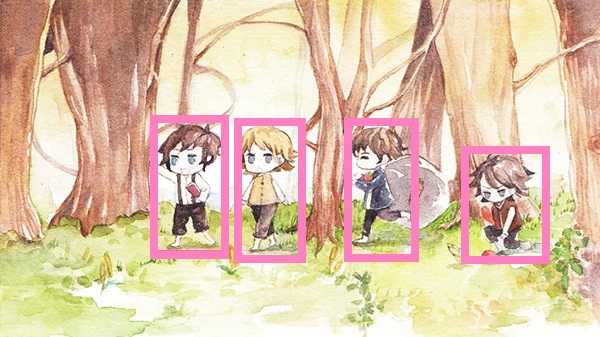}
  \end{minipage}
  }
  \hspace{0.01in}
  \subfigure[\footnotesize SW baseline]{
  \begin{minipage}[t]{0.18\linewidth}
    \includegraphics[width=1.3in]{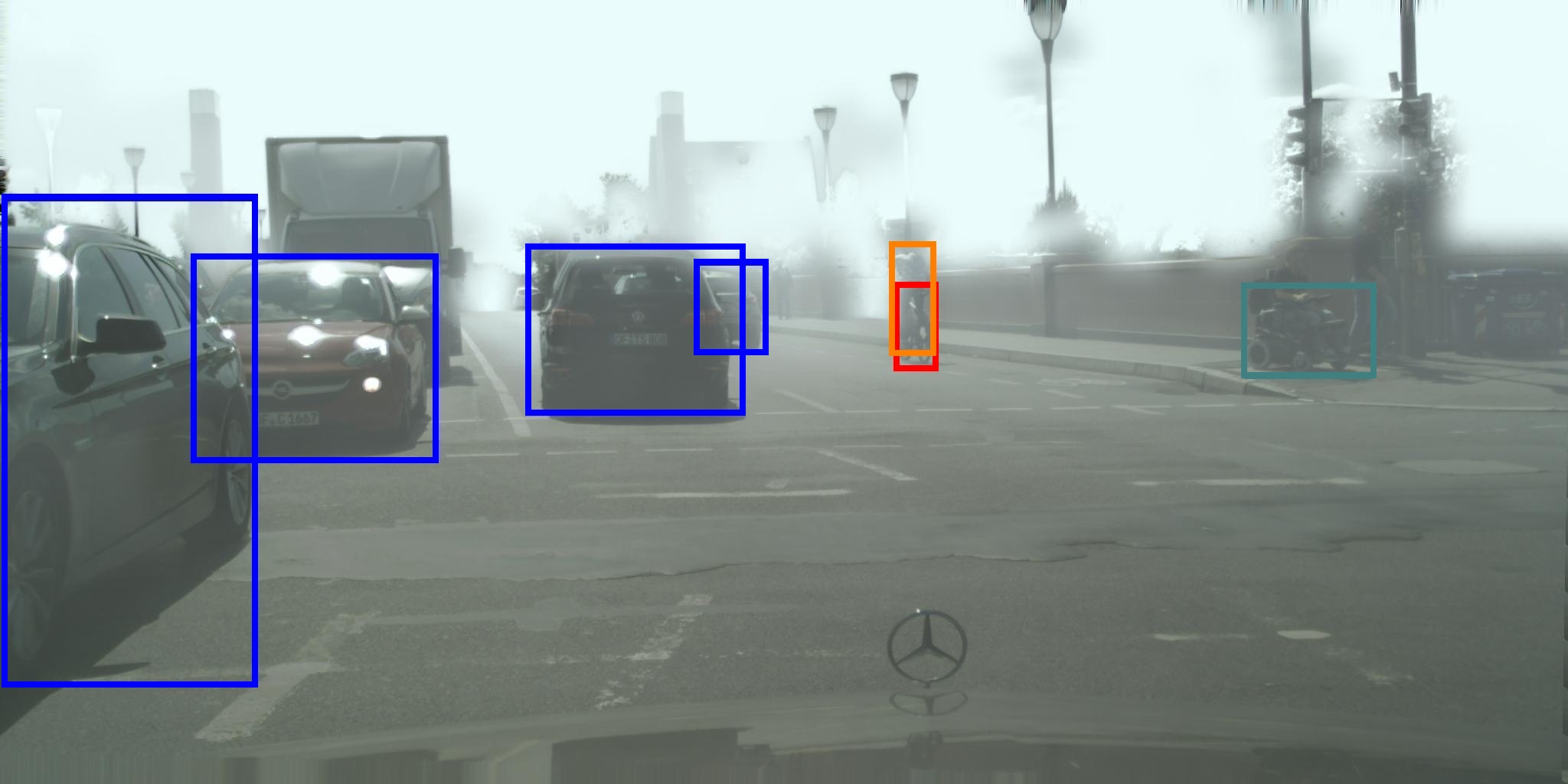}\\ \vspace{-0.1in}
    \includegraphics[width=1.3in]{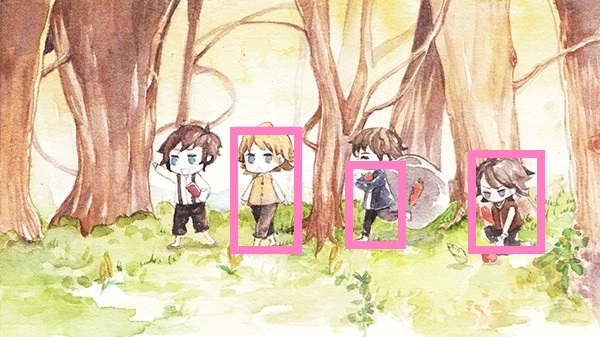}
  \end{minipage}
  }
  \hspace{0.01in}
  \subfigure[\footnotesize One Step Training]{
  \begin{minipage}[t]{0.18\linewidth}
    \includegraphics[width=1.3in]{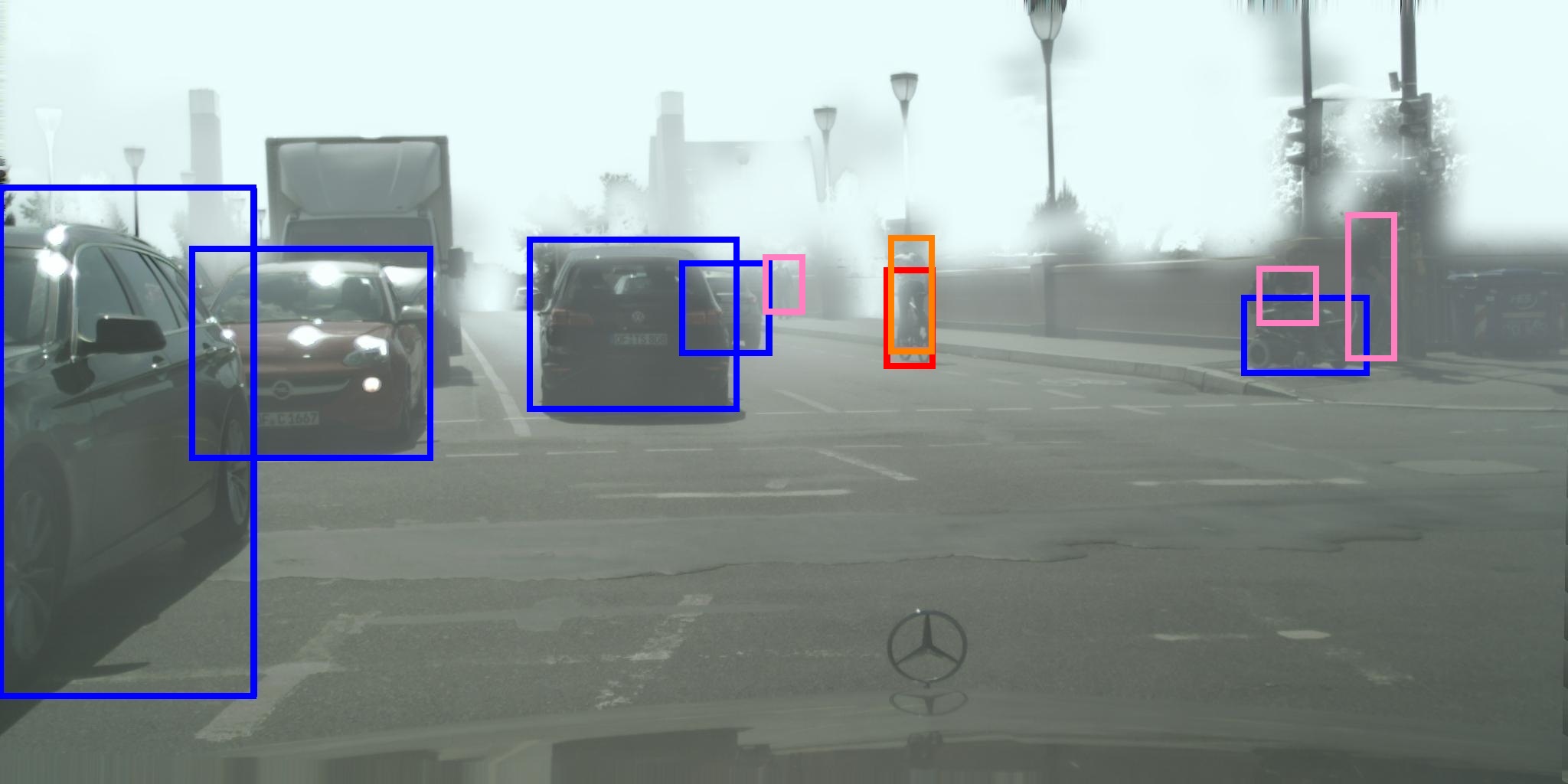}\\ \vspace{-0.1in}
    \includegraphics[width=1.3in]{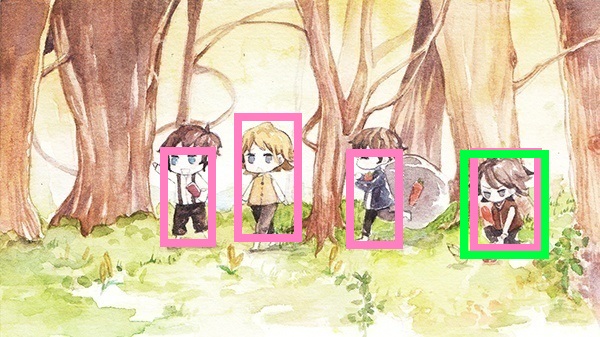}
  \end{minipage}
  }
  \hspace{0.01in}
  \subfigure[\footnotesize Two-Step Training]{
  \begin{minipage}[t]{0.18\linewidth}
    \includegraphics[width=1.3in]{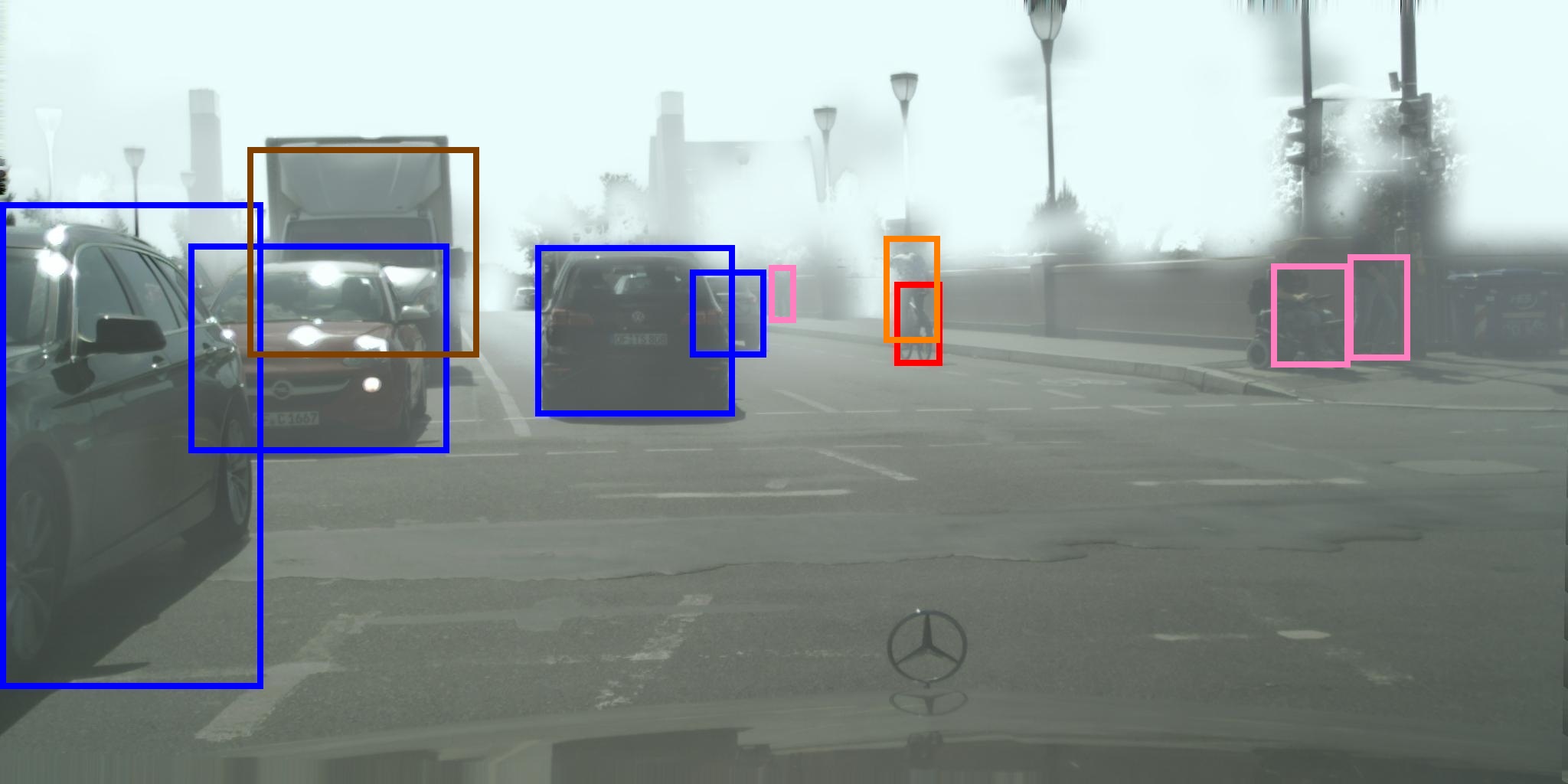}\\ \vspace{-0.1in}
    \includegraphics[width=1.3in]{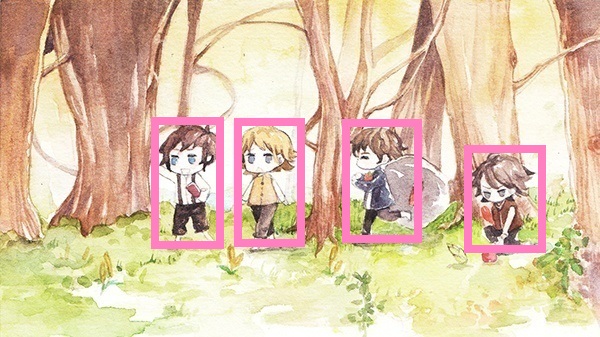}
  \end{minipage}
  }
\end{center}
\vspace{-0.1in}
\caption{Detection results on the FoggyCityscapes and Watercolor scene. `GT' indicates the groundtruth result. `One-Step Training' denotes we integrate all loss functions and use one optimization step to train SW-VDD. We can see that using two training steps could localize and recognize objects existing in the two images accurately, e.g., the \textcolor[rgb]{0.5,0,0}{truck}, \textcolor[rgb]{0,0,1}{car}, \textcolor[rgb]{1.0,0.5,0.75}{person}, \textcolor[rgb]{1,0,0}{bicycle}, and \textcolor[rgb]{0,0.8,0}{bird}.}
\label{fig6}
\vspace{-0.1in}
\end{figure*}

\subsection{Result Analysis of Single-target DAOD}

\textbf{Results on FoggyCityscapes.} Table \ref{table1} shows the results of FoggyCityscapes. Here, VGG16 \cite{VGG} is taken as the backbone. Through plugging our disentanglement into domain adaptive Faster R-CNN methods, the performance can be improved significantly. Particularly, for SW \cite{saito2019strong} and ICCR \cite{xu2020exploring}, our method separately improves the performance by 3.6\% and 2.6\%. This demonstrates decomposing domain-invariant features is helpful for alleviating the domain-shift impact on object detection. 

\begin{table}
\centering
\small
\scalebox{0.95}{
\tabcolsep=3.5pt
\begin{tabular}{l|cccccc|c}
\toprule[1.5pt]
Method   & bike & bird & car  & cat  & dog  & person & mAP  \\ \hline
Source Only &68.8 & 46.8 & 37.2 & 32.7 & 21.3 & 60.7   & 44.6 \\ \hline
BDC-Faster \cite{saito2019strong} &68.6 & 48.3 & 47.2 & 26.5 & 21.7 & 60.5 & 45.5 \\ \hline
DAF \cite{chen2018domain} &75.2 & 40.6 & 48.0 & 31.5 & 20.6 & 60.0 & 46.0\\ \hline
WST-BSR \cite{kim2019self} & 75.6 & 45.8 & 49.3 & 34.1 & 30.3 & 64.1 & 49.9 \\ \hline
MAF \cite{he2019multi} & 73.4 & 55.7 & 46.4 & 36.8 & 28.9 & 60.8 & 50.3 \\ \hline
DC \cite{liu2020domain} & 76.7 & 53.2 & 45.3 & {\bf 41.6} & 35.5 & {\bf 70.0} & 53.7 \\ \hline
ATF \cite{he2020domain} & 78.8 & {\bf 59.9} & 47.9 & 41.0 & 34.8 & 66.9 & 54.9 \\ \hline
SCL \cite{shen2019scl} & 82.2 & 55.1 & {\bf 51.8} & 39.6 & 38.4 & 64.0 & 55.2 \\ \hline
MCAR \cite{zhao2020adaptive} & 87.9 & 52.1 & {\bf 51.8} & {\bf 41.6} & 33.8 & 68.8 & 56.0 \\ \hline \hline
SW \cite{saito2019strong} & 82.3 & 55.9 & 46.5 & 32.7 & 35.5 & 66.7 & 53.3 \\
SW-VDD (ours) & {\bf 90.0} & 56.6 & 49.2 & 39.5 & {\bf 38.8} & 65.3 & {\bf 56.6} \\
\bottomrule[1.5pt]
\end{tabular}}
\caption{Results (\%) on adaptation from Pascal to Watercolor.}\label{table2}
\vspace{-0.15in}
\end{table}

The first row of Fig. \ref{fig6} shows one detection example from the FoggyCityscapes dataset. Here, we take SW \cite{saito2019strong} as an example. We can see that compared with SW, our method localizes and recognizes objects existing in the foggy image accurately. This further shows our method is effective.

\textbf{Results on Watercolor.} Table \ref{table2} shows the Watercolor results. Here, we use ResNet101 \cite{ResNet} as the backbone. We can see plugging vector-decomposed disentanglement into SW \cite{saito2019strong} improves its performance significantly. Besides, MCAR \cite{zhao2020adaptive} exploits multi-label object recognition as a dual auxiliary task to improve the alignment. We can see that our method outperforms MCAR by 0.6\%. These all demonstrate our method is effective. And our method could alleviate the impact of the watercolor style.

The second row of Fig. \ref{fig6} shows one watercolor example. We can see compared with SW, our method could localize and recognize objects accurately. These further demonstrate employing vector-decomposed disentanglement could indeed alleviate the domain-shift impact.

\begin{table}[ht]
\centering
\small
\scalebox{0.85}{
\tabcolsep=4.2pt
\begin{tabular}{l|ccccccc|c}
\toprule[1.5pt]
Method   & bus & bike & car  & motor  & person  & rider & truck & mAP  \\ \hline
Source Only &38.6 & 21.5 & 51.7 & 12.0 & 19.7 & 13.6 & 40.9  & 28.3 \\ \hline
CT \cite{zhao2020collaborative} & 35.5 & 20.3 & 50.9 & 7.9 & 21.6 & 16.1 & 34.4 & 26.7 \\ \hline
SCL \cite{shen2019scl} & 34.8 & 19.2 & 50.8 & 13.2 & 25.9 & 18.0 & 38.1 & 28.6 \\ \hline
HTCN \cite{chen2020harmonizing} &35.9 & 21.1 & 51.1 & 13.7 & 24.0 & 16.6 & 39.0 & 28.8 \\ \hline
DAF \cite{chen2018domain} & 43.6 & 27.5 & 52.3 & 16.1 & 28.5 & 21.7 & 44.8 & 33.5 \\ \hline \hline
SW \cite{saito2019strong} & 40.0 & 22.8 & 51.4 & 15.4 & 26.3 & 20.3 & 44.2 & 31.5 \\
SW-VDD & 46.1 & 31.1 & 54.4 & 25.3 & {\bf 31.0} & 22.4 & 47.6 & 36.9 \\
ICCR \cite{xu2020exploring} & 43.8 & 28.5 & 52.4 & 22.7 & 29.2 & 21.9 & 45.6 & 34.9 \\
ICCR-VDD & {\bf 47.9} & {\bf 33.2} & {\bf 55.1} & {\bf 26.1} & 30.5 & {\bf 23.8} & {\bf 48.1} & {\bf 37.8} \\
\bottomrule[1.5pt]
\end{tabular}}
\caption{Results (\%) on adaptation from Daytime-sunny to Dusk-rainy. Here, we directly run the released codes of the compared methods to obtain the results.}\label{table3}
\vspace{-0.15in}
\end{table}

\textbf{Results on Dusk-rainy.} Table \ref{table3} shows the results of Daytime-sunny $\rightarrow$ Dusk-rainy. ResNet101 \cite{ResNet} is taken as the backbone. We can see that for this scene, the adaptation performance of state-of-the-art methods, e.g., CT \cite{zhao2020collaborative} and HTCN \cite{chen2020harmonizing}, is weak. Besides, we can also see that plugging the disentanglement into SW \cite{saito2019strong} and ICCR \cite{xu2020exploring} improves their performance significantly. The performance is separately improved by 5.4\% and 2.9\%. This further demonstrates vector-decomposed disentanglement is capable of disentangling domain-invariant features, which is helpful for alleviating the domain-shift impact on object detection.

\begin{figure*}[ht]
\begin{center}
  \subfigure{
  \begin{minipage}[t]{0.3\linewidth}
    \includegraphics[width=2.2in,height=1.2in]{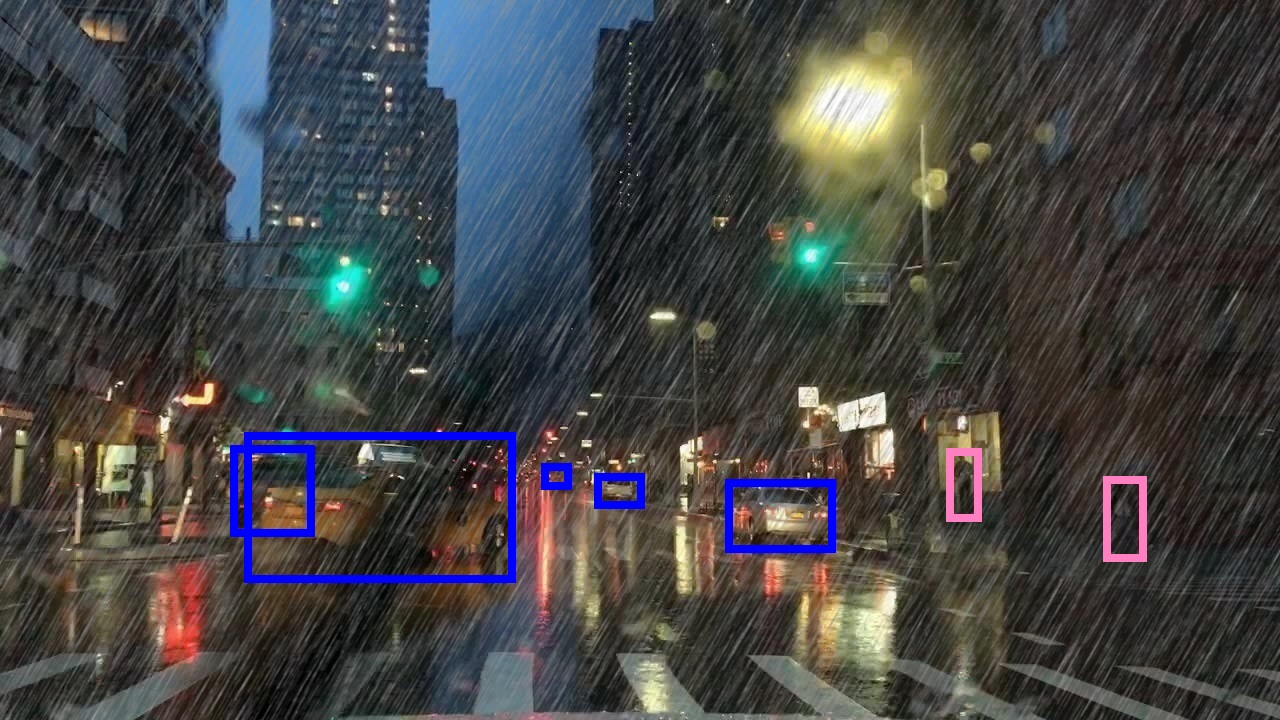} \vspace{-0.1in} \\
    \includegraphics[width=2.2in,height=1.2in]{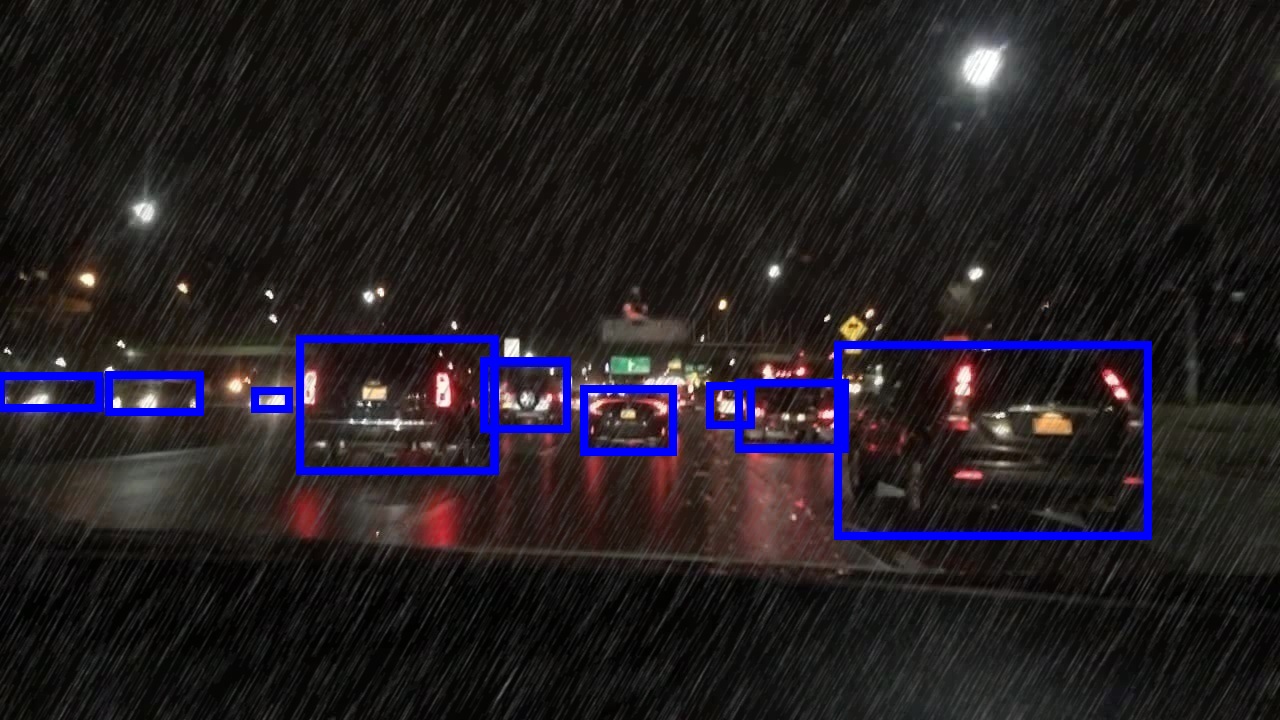}
  \end{minipage}
  }
  \hspace{0.1in}
  \subfigure{
  \begin{minipage}[t]{0.3\linewidth}
    \includegraphics[width=2.2in,height=1.2in]{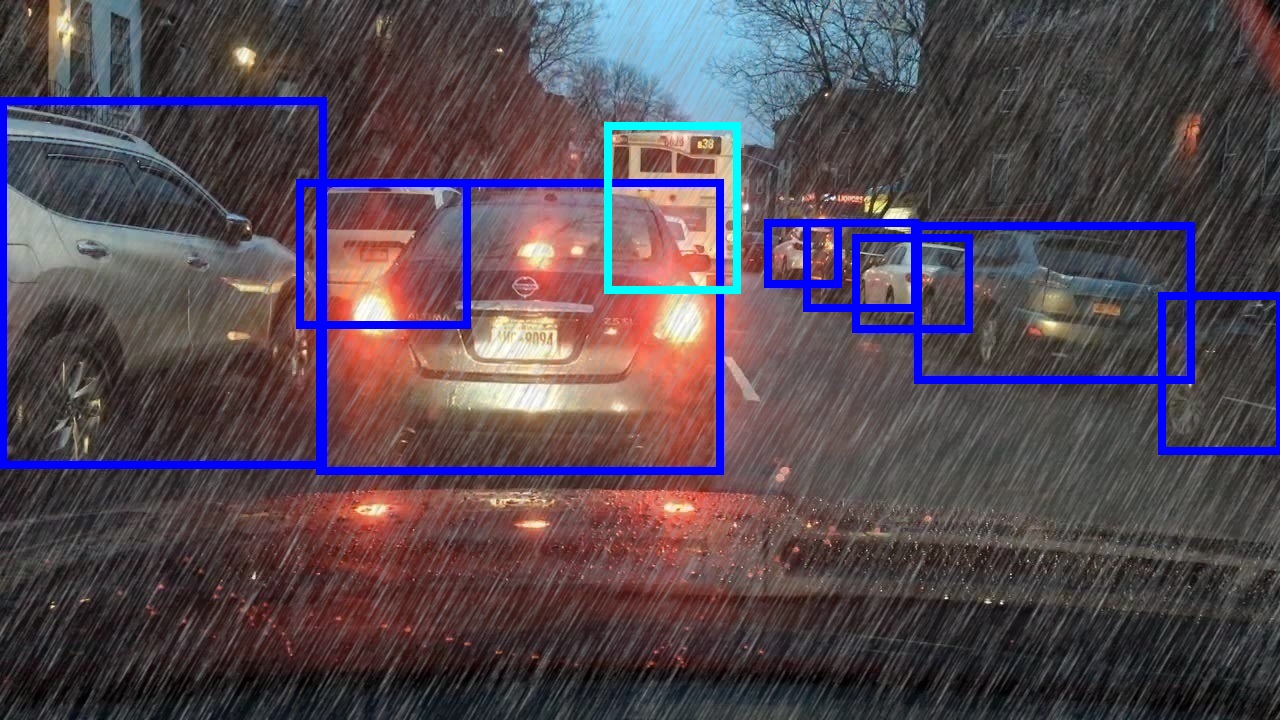} \vspace{-0.1in} \\
    \includegraphics[width=2.2in,height=1.2in]{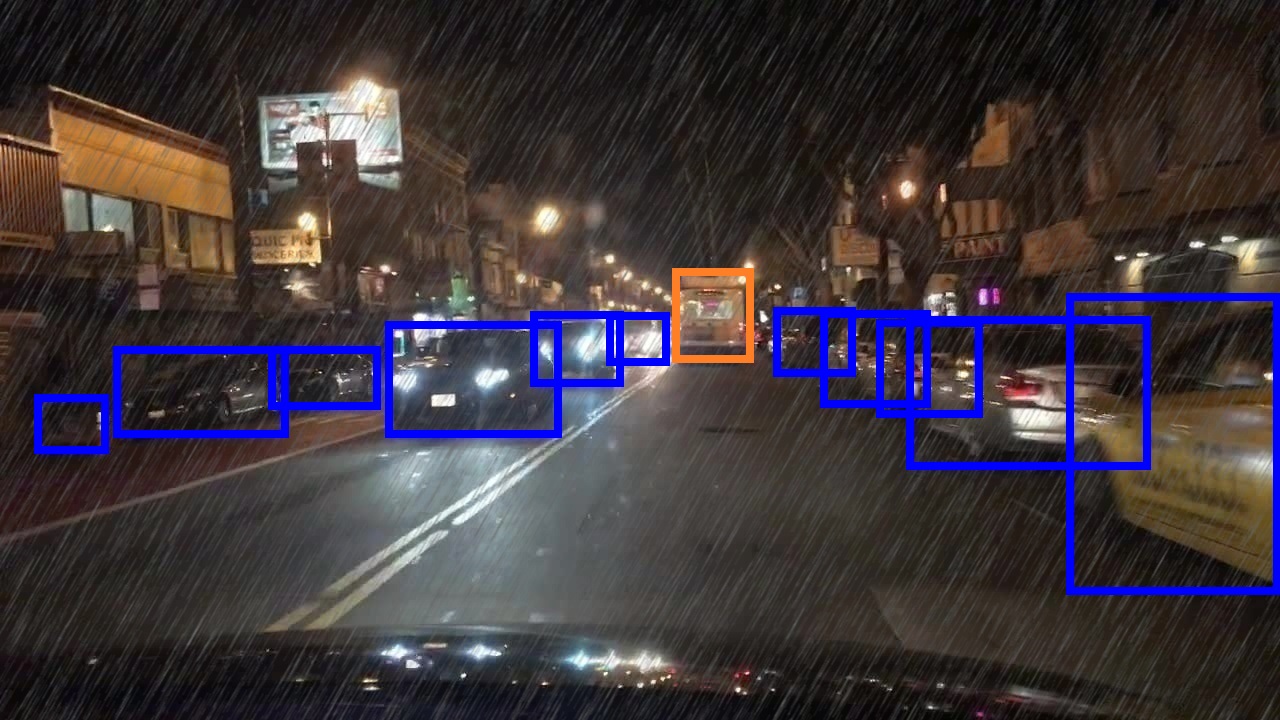}
  \end{minipage}
  }
  \hspace{0.1in}
  \subfigure{
  \begin{minipage}[t]{0.3\linewidth}
    \includegraphics[width=2.2in,height=1.2in]{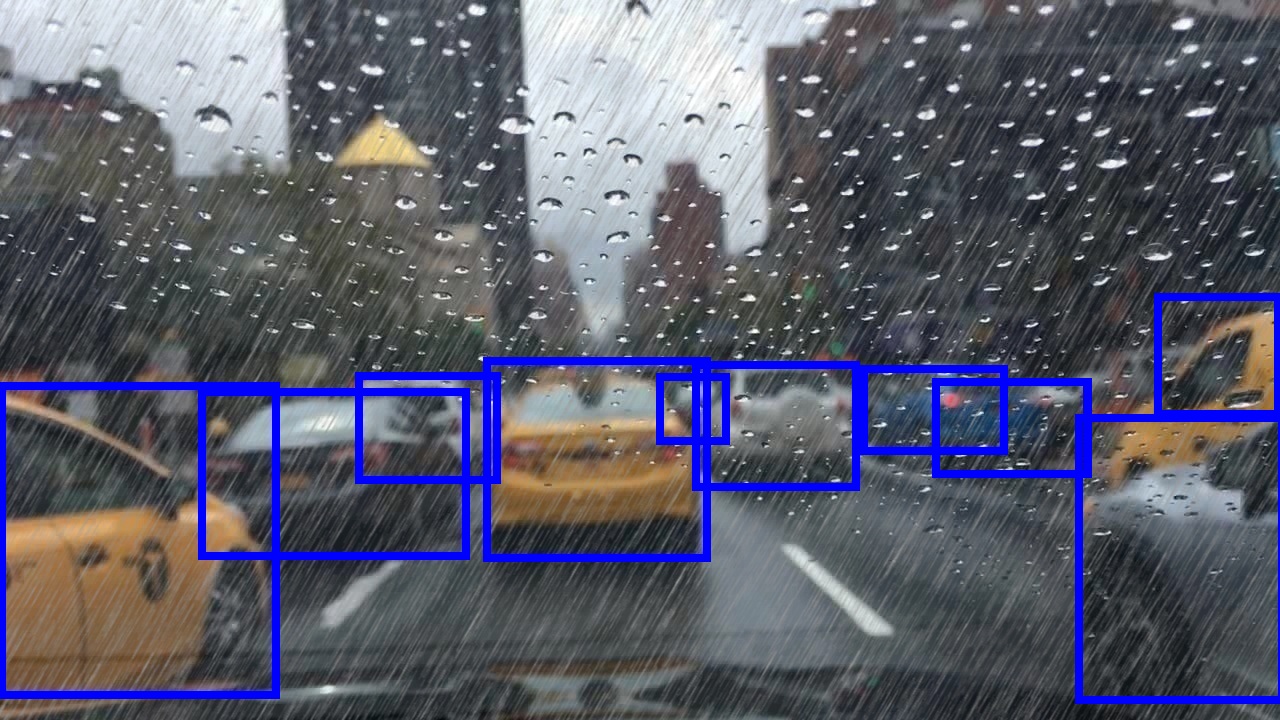} \vspace{-0.1in} \\
    \includegraphics[width=2.2in,height=1.2in]{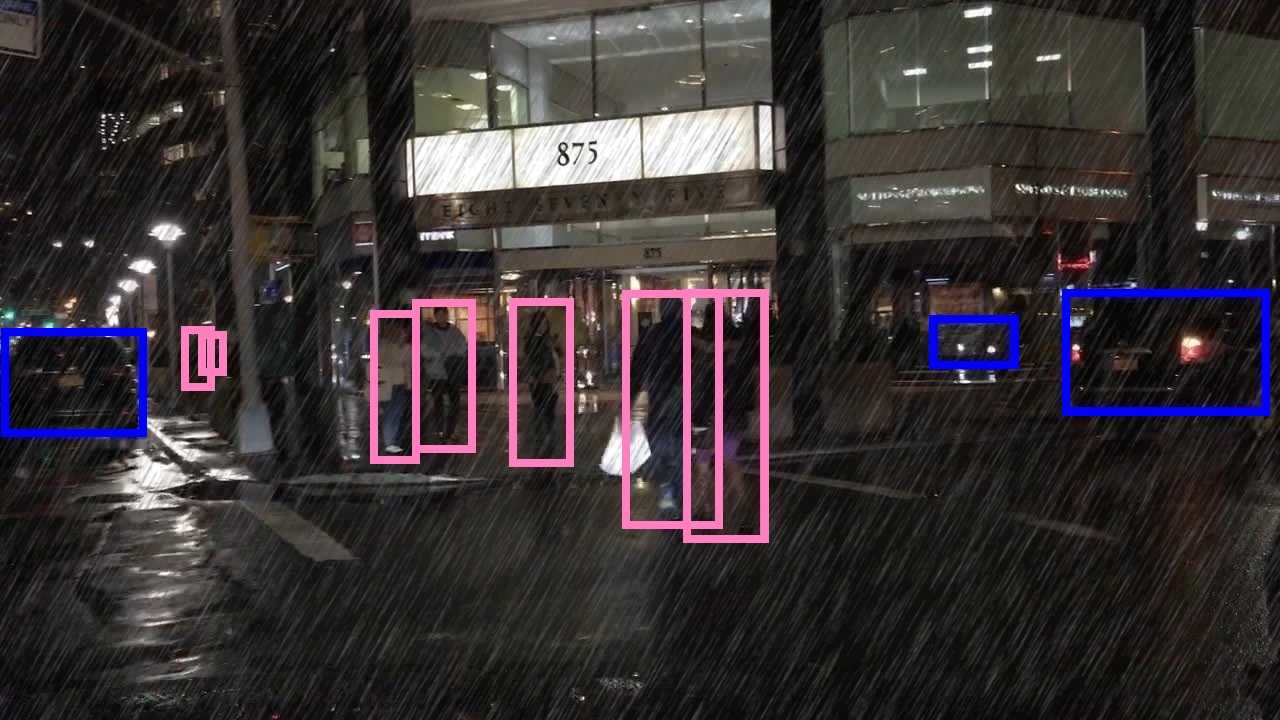}
  \end{minipage}
  }
\end{center}
\vspace{-0.1in}
\caption{The first and second row separately show the detection results on the ``Daytime-sunny $\rightarrow$ Dusk-rainy'' and ``Daytime-sunny $\rightarrow$ Night-rainy''. We can see our method detects objects existing in these images, which shows the effectiveness of our method.}
\label{fig7}
\vspace{-0.1in}
\end{figure*}

The first row of Fig. \ref{fig7} shows three detection examples of the dusk-rainy scene. We can see that this is a challenging adaptation scene. The images are very obscure. Our method localizes and recognizes objects existing in these images accurately, which further demonstrates the effectiveness of vector-decomposed disentanglement.

\textbf{Results on Night-rainy.} Table \ref{table4} shows the results of Daytime-sunny $\rightarrow$ Night-rainy. ResNet101 \cite{ResNet} is taken as the backbone. We can see that for this scene, the performance of many adaptation methods \cite{chen2020harmonizing,shen2019scl,zhao2020collaborative} is weak. For example, the mAP value of HTCN and CT is lower than 20\%. Plugging the disentanglement into SW \cite{saito2019strong} and ICCR \cite{xu2020exploring} improves their performance significantly. The performance is improved by 5.7\% and 3.1\%. Particularly, for each object category, our method outperforms SW \cite{saito2019strong} and ICCR \cite{xu2020exploring}. This further demonstrates the effectiveness of vector-decomposed disentanglement.

\begin{table}[ht]
\centering
\small
\scalebox{0.85}{
\tabcolsep=4.2pt
\begin{tabular}{l|ccccccc|c}
\toprule[1.5pt]
Method   & bus & bike & car  & motor  & person  & rider & truck & mAP  \\ \hline
Source Only &23.4 & 13.3 & 31.8 & 1.5 & 10.2 & 10.9 & 23.2  & 16.3 \\ \hline
CT \cite{zhao2020collaborative} & 22.4 & 9.7 & 27.4 & 0.6 & 9.3 & 9.3 & 13.4 & 13.1 \\ \hline
SCL \cite{shen2019scl} & 20.0 & 9.2 & 33.2 & 0.3 & 11.9 & 10.6 & 26.4 & 15.9 \\ \hline
HTCN \cite{chen2020harmonizing} &22.8 & 9.4 & 30.7 & 0.7 & 11.9 & 4.8 & 22.0 & 14.6 \\ \hline
DAF \cite{chen2018domain} & 23.8 & 12.0 & 37.7 & 0.2 & 14.9 & 4.0 & 29.0 & 17.4 \\ \hline \hline
SW \cite{saito2019strong} & 24.7 & 10.0 & 33.7 & 0.6 & 13.5 & 10.4 & 29.1 & 17.4 \\
SW-VDD & 31.7 & 15.3 & 38.0 & {\bf 11.1} & 18.2 & 16.7 & {\bf 30.8} & 23.1 \\
ICCR \cite{xu2020exploring} & 32.5 & 12.1 & 36.2 & 1.3 & 16.1 & 17.0 & 29.3 & 20.6 \\
ICCR-VDD & {\bf 34.8} & {\bf 15.6} & {\bf 38.6} & 10.5 & {\bf 18.7} & {\bf 17.3} & 30.6 & {\bf 23.7} \\
\bottomrule[1.5pt]
\end{tabular}}
\caption{Results (\%) on Daytime-sunny $\rightarrow$ Night-rainy.}\label{table4}
\vspace{-0.15in}
\end{table}

The second row of Fig. \ref{fig7} shows three detection examples of the night-rainy scene. We can see for this scene, the brightness of images is very low. Meanwhile, the rainy images are very obscure. Our method localizes and recognizes objects existing in the night-rainy images accurately. This demonstrates extracting domain-invariant representations is helpful for alleviating the domain-shift impact. Our method could extract domain-invariant representations effectively.

\subsection{Ablation Analysis}

Based on the single-target case, we plug our method into SW \cite{saito2019strong} to make an ablation analysis. Table \ref{ablation} shows the results. We can see that for our model, employing two training steps is effective. Particularly, two-step training outperforms one-step training by 3.4\% and 2.1\%. This shows our optimization mechanism promotes the model to extract domain-invariant representations, which is beneficial for DAOD. In Fig. \ref{fig6}(d), we show two examples based on one training step. We can see using two training steps could detect objects existing in the two images accurately. Moreover, we can also see that the orthogonal loss could improve the performance significantly. This shows the orthogonal loss is indeed helpful for promoting DIR and DSR to be independent, which improves the disentangled ability.

\begin{table}
\centering
\small
\scalebox{0.95}{
\begin{tabular}{c|ccc|c|c}
\toprule[1.5pt]
Method  & One-step & Two-step & OL & C $\rightarrow$ F & V $\rightarrow$ W \\ \hline
SW-VDD &$\checkmark$ & & & 33.2\% & 52.7\% \\
SW-VDD &$\checkmark$ & &$\checkmark$ & 34.5\% & 54.5\% \\
SW-VDD & &$\checkmark$ & &36.5\% & 54.9\% \\
SW-VDD & &$\checkmark$ &$\checkmark$ & \textbf{37.9\%} & \textbf{56.6\%} \\ \bottomrule[1.5pt]
\end{tabular}}
\caption{Ablation analysis of our method. Here, we use mAP as the metric. `One-step' and `Two-step' indicate we use one training step and two training steps to optimize our model, respectively. `OL' denotes the orthogonal loss. `${\rm C} \rightarrow {\rm F}$' denotes the adaptation from Cityscapes to FoggyCityscapes and employs VGG16 as the backbone. `${\rm V} \rightarrow {\rm W}$' denotes the adaptation from VOC to Watercolor and utilizes ResNet101 as the backbone.}\label{ablation}
\vspace{-0.15in}
\end{table}

\textbf{Compared with traditional disentanglement.} To further demonstrate the effectiveness of our method, we replace our method with the traditional disentanglement \cite{peng2019domain,wu2021instance}. Other components are kept unchanged. We employ the same training steps to optimize the model. Based on FoggyCityscapes and Watercolor dataset, the adaptation performance of the traditional disentanglement is 34.1\% and 54.6\%, which is weaker than our method. Besides, since our method does not include the reconstruction stage, our method owns much fewer parameters and computational costs. These all demonstrate the performance of our method outperforms the traditional disentangled method. Meanwhile, this also shows that our vector-decomposed disentanglement could extract domain-invariant features effectively, which improves the detection performance.

\begin{figure*}
\vspace{-0.1em}
\begin{center}
  \subfigure[\footnotesize GT]{
  \begin{minipage}[t]{0.12\linewidth}
    \includegraphics[width=0.9in,height=1.1in]{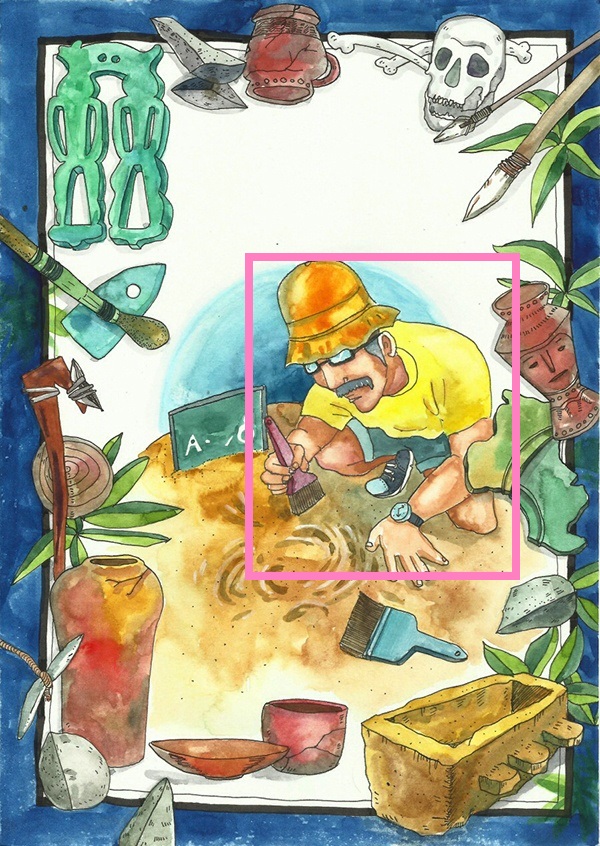} \\ \vspace{-0.1in}
    \includegraphics[width=0.9in,height=1.1in]{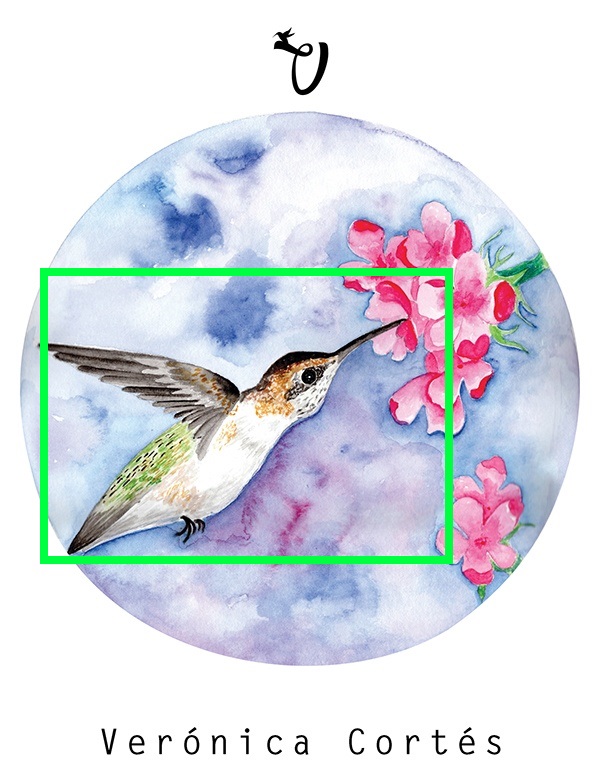}  \\ \vspace{-0.1in}
    \includegraphics[width=0.9in,height=1.0in]{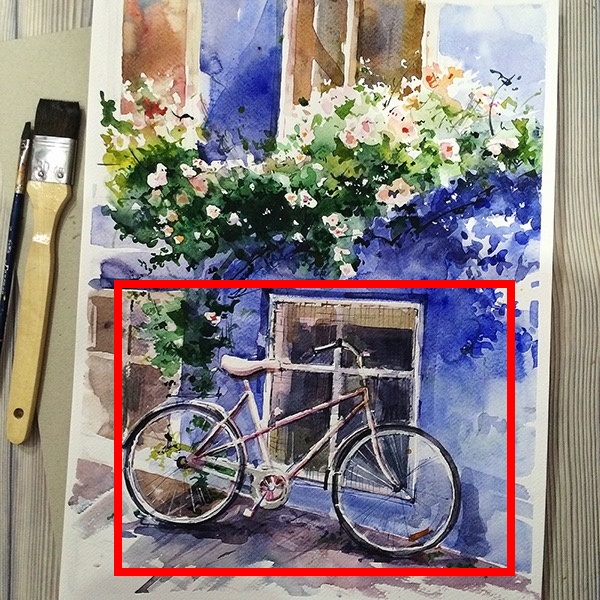}
  \end{minipage}
  }
  \hspace{0.02cm}
  \subfigure[\footnotesize TD-Results]{
  \begin{minipage}[t]{0.12\linewidth}
    \includegraphics[width=0.9in,height=1.1in]{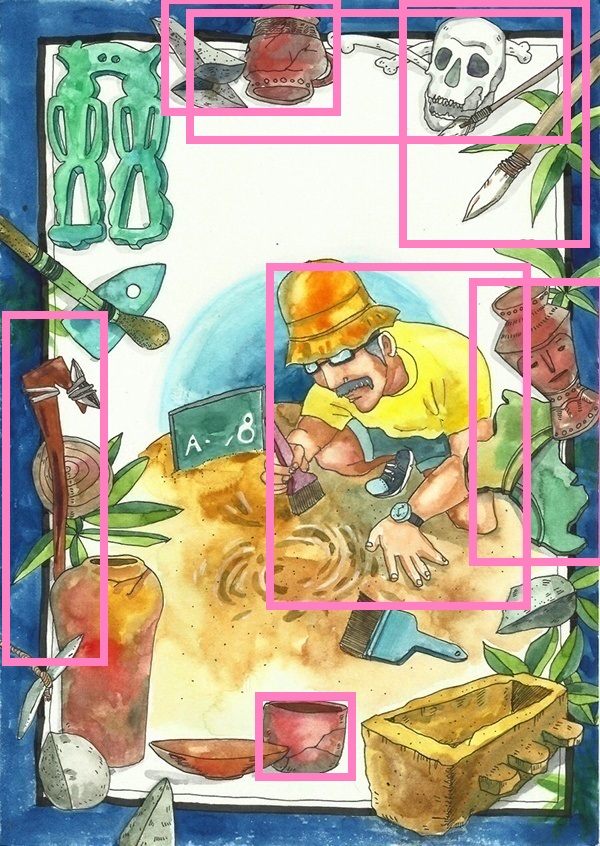} \\ \vspace{-0.1in}
    \includegraphics[width=0.9in,height=1.1in]{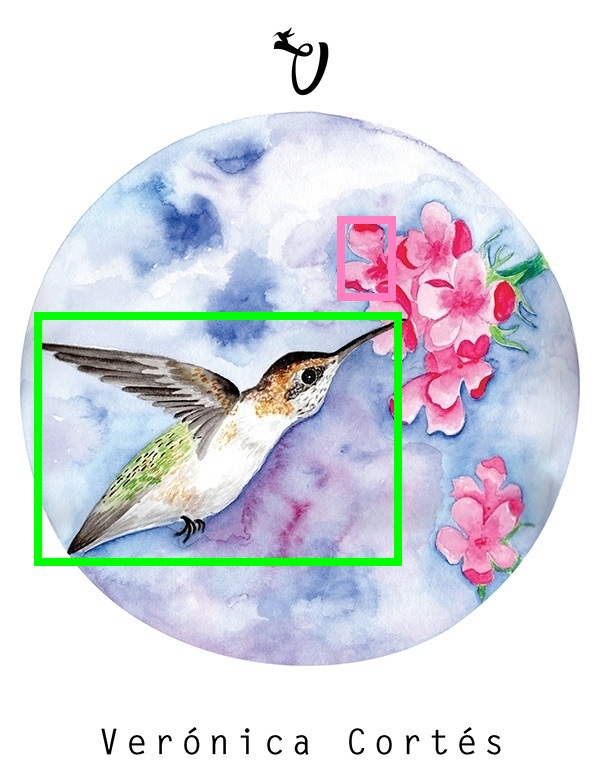} \\ \vspace{-0.1in}
    \includegraphics[width=0.9in,height=1.0in]{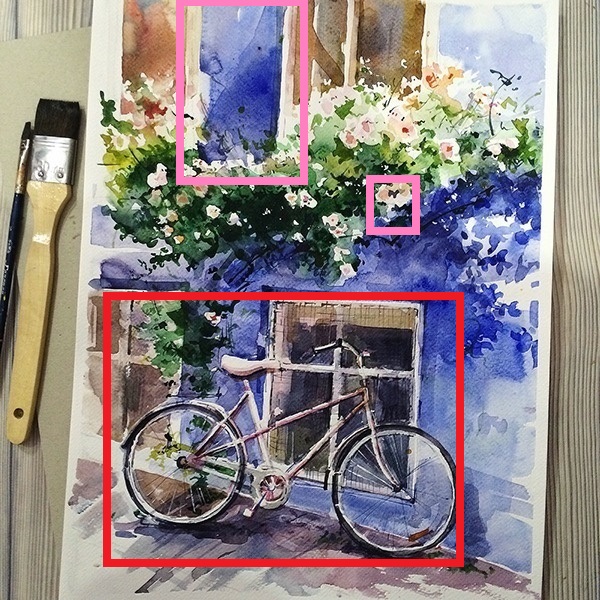}
  \end{minipage}
  }
  \hspace{-0.01em}
  \subfigure[\footnotesize VDD-Results]{
  \begin{minipage}[t]{0.12\linewidth}
    \includegraphics[width=0.9in,height=1.1in]{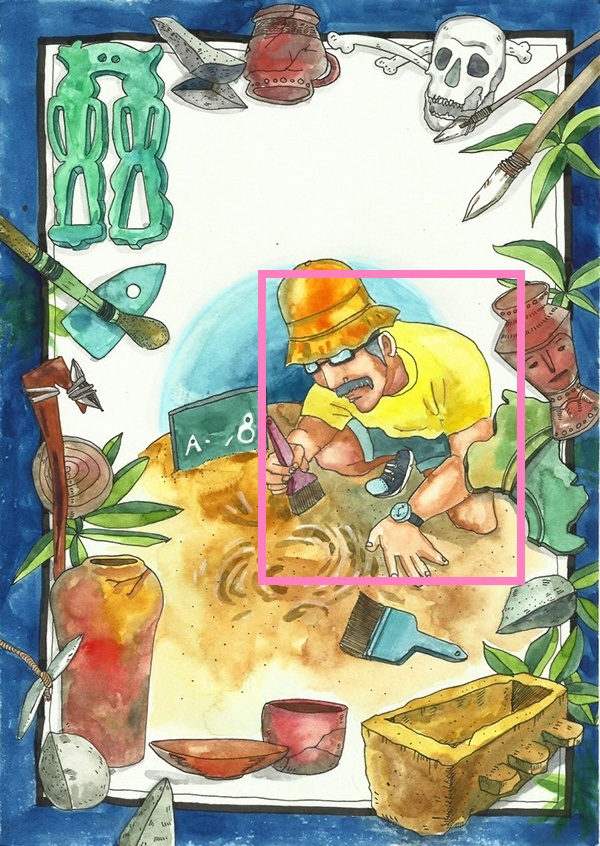} \\ \vspace{-0.1in}
    \includegraphics[width=0.9in,height=1.1in]{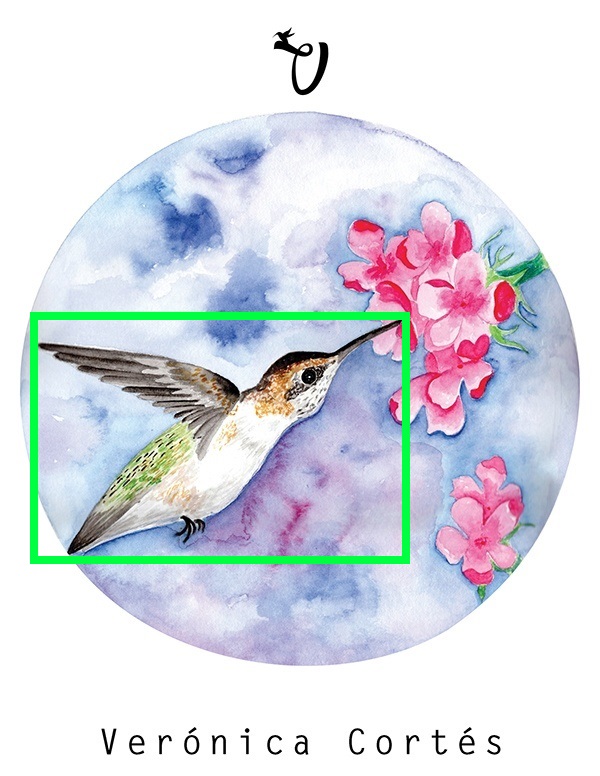}  \\ \vspace{-0.1in}
    \includegraphics[width=0.9in,height=1.0in]{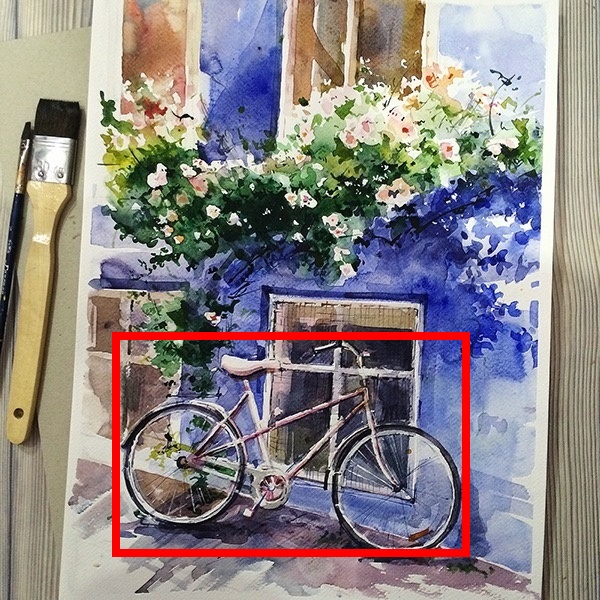}
  \end{minipage}
  }
  \hspace{0.04cm}
  \subfigure[\footnotesize TD-Base]{
  \begin{minipage}[t]{0.12\linewidth}
    \includegraphics[width=0.9in,height=1.1in]{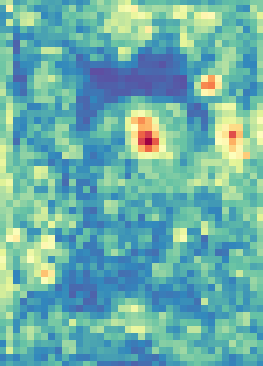} \\ \vspace{-0.1in}
    \includegraphics[width=0.9in,height=1.1in]{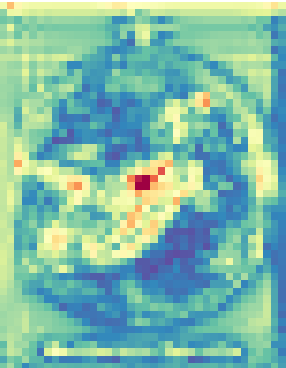} \\ \vspace{-0.1in}
    \includegraphics[width=0.9in,height=1.0in]{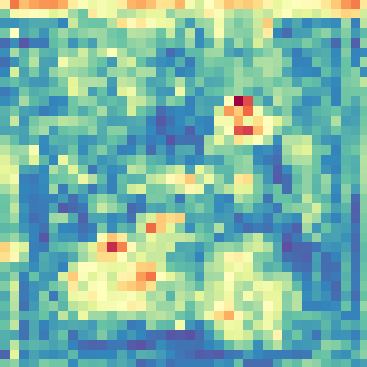}
  \end{minipage}
  }
  \hspace{-0.01em}
  \subfigure[\footnotesize VDD-Base]{
  \begin{minipage}[t]{0.12\linewidth}
    \includegraphics[width=0.9in,height=1.1in]{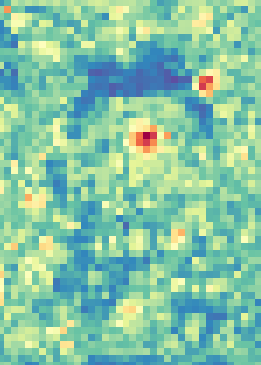} \\ \vspace{-0.1in}
    \includegraphics[width=0.9in,height=1.1in]{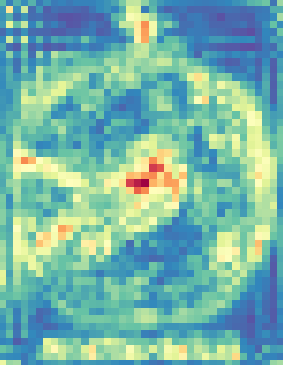}  \\ \vspace{-0.1in}
    \includegraphics[width=0.9in,height=1.0in]{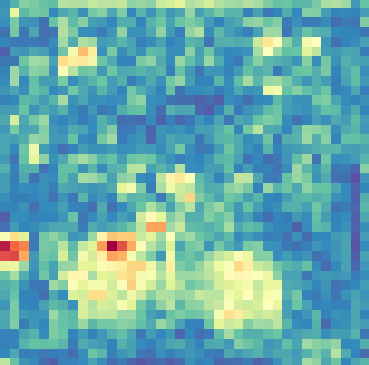}
  \end{minipage}
  }
  \hspace{0.04cm}
  \subfigure[\footnotesize TD-DIR]{
  \begin{minipage}[t]{0.12\linewidth}
    \includegraphics[width=0.9in,height=1.1in]{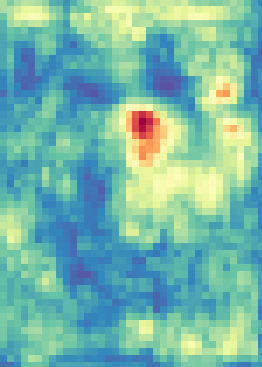} \\ \vspace{-0.1in}
    \includegraphics[width=0.9in,height=1.1in]{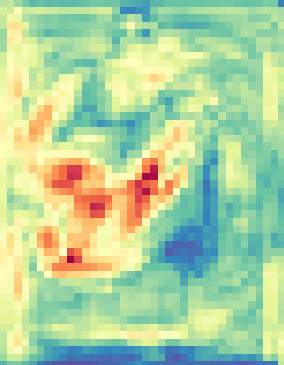} \\ \vspace{-0.1in}
    \includegraphics[width=0.9in,height=1.0in]{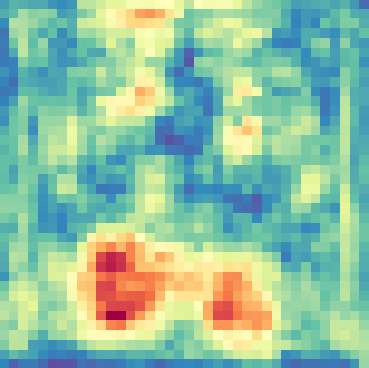}
  \end{minipage}
  }
  \hspace{-0.01em}
  \subfigure[\footnotesize VDD-DIR]{
  \begin{minipage}[t]{0.12\linewidth}
   \includegraphics[width=0.9in,height=1.1in]{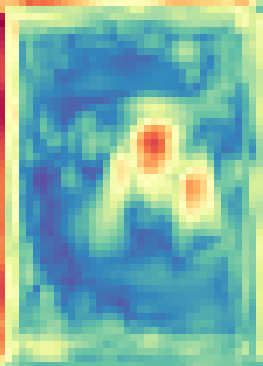} \\ \vspace{-0.1in}
   \includegraphics[width=0.9in,height=1.1in]{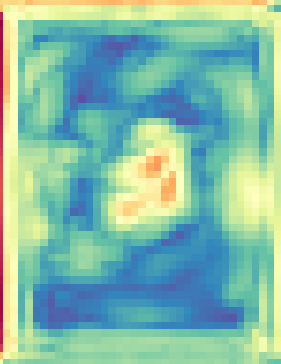}  \\ \vspace{-0.1in}
   \includegraphics[width=0.9in,height=1.0in]{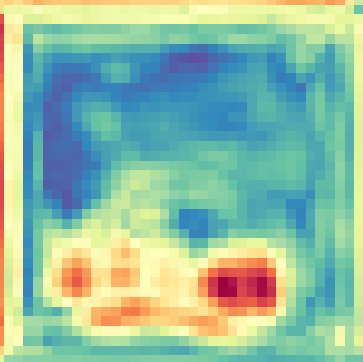}
  \end{minipage}
  }
\end{center}
\vspace{-0.1in}
\caption{Comparisons of feature maps extracted by our vector-decomposed disentanglement (VDD) and traditional disentanglement (TD). Here, `TD-Results' indicates detection results of TD. `TD-Base' and `VDD-Base' separately indicate the feature map used for disentanglement. `TD-DIR' and `VDD-DIR' separately indicate the DIR extracted by TD and VDD. These examples are from the `Pascal VOC $\rightarrow$ Watercolor' scene. For each feature map, the channels corresponding to the maximum value are selected for visualization.}
\label{fig8}
\vspace{-0.1in}
\end{figure*}

\textbf{Visualization analysis.} In Fig. \ref{fig8}, we compare DIR extracted by our disentangled method and traditional disentanglement. We find that compared with traditional disentanglement, the DIR extracted by our vector-decomposed disentanglement contains much less domain-specific information. Particularly, for these examples, we can see that the DIR extracted by the traditional disentanglement contains much more domain-specific information, e.g., the TD-DIR (Fig. \ref{fig8}(f)) of the bird image, which leads to the incorrect detections. This further demonstrates the effectiveness of our vector-decomposed disentanglement.

\begin{table}[ht]
\centering
\small
\scalebox{0.85}{
\tabcolsep=4.2pt
\begin{tabular}{l|ccccccc|c}
\toprule[1.5pt]
Method   & bus & bike & car  & motor  & person  & rider & truck & mAP  \\ \hline
Source Only   &35.1    &19.3    &44.0   &8.8     &17.5      &12.8     &37.7     &25.0 \\ \hline
DAF \cite{chen2018domain}     &35.9    &18.3    &44.2   &10.1    &22.0      &17.9     &39.9     &26.9 \\ \hline
CT \cite{zhao2020collaborative}      &31.3    &15.4    &41.7   &8.4     &19.1      &15.3     &32.3     &23.4 \\ \hline
SCL \cite{shen2019scl}     &32.7    &19.7    &44.9   &10.5    &22.9      &18.5     &38.3     &26.8 \\ \hline
SW \cite{saito2019strong}      &36.9    &20.7    &45.1   &6.6    &23.1      &16.9     &41.5     &27.3 \\ \hline
ICCR  \cite{xu2020exploring}   &38.8    &20.4    &44.6   &11.7    &24.7      &15.4     &41.6     &28.2 \\ \midrule
SW-VDD      &\textbf{41.8}    &\textbf{26.8}    &\textbf{48.6}   &\textbf{17.9}    &\textbf{27.0}      &\textbf{22.2}     &\textbf{44.1}     &\textbf{32.6} \\
\bottomrule[1.5pt]
\end{tabular}}
\caption{Results (\%) on the compound target domain.}\label{table6}
\vspace{-0.15in}
\end{table}
\begin{table}[ht]
\centering
\small
\scalebox{0.85}{
\tabcolsep=4.2pt
\begin{tabular}{l|ccccccc|c}
\toprule[1.5pt]
Method   & bus & bike & car  & motor  & person  & rider & truck & mAP  \\ \hline
Source Only   &38.6    &21.5    &51.7   &12.0    &19.7      &13.6     &40.9     &28.3 \\ \hline
DAF \cite{chen2018domain}     &39.5    &21.0    &51.6   &12.6    &24.8      &20.5     &42.7     &30.4 \\ \hline
CT \cite{zhao2020collaborative}      &34.9    &17.6    &49.8   &11.6    &21.9      &17.9     &35.6     &27.0 \\ \hline
SCL \cite{shen2019scl}     &35.7    &22.3    &50.7   &14.8    &25.3      &19.9     &40.1     &29.8 \\ \hline
SW \cite{saito2019strong}      &39.2    &24.6    &49.6   &9.2     &25.5      &19.3     &43.7     &30.1 \\ \hline
ICCR  \cite{xu2020exploring}   &42.0    &21.9    &51.5   &16.5    &27.2      &16.8     &44.1     &31.4 \\ \midrule
SW-VDD      &\textbf{43.7}    &\textbf{30.3}    &\textbf{52.7}   &\textbf{22.3}    &\textbf{29.7}      &\textbf{24.8}     &\textbf{46.4}     &\textbf{35.7} \\
\bottomrule[1.5pt]
\end{tabular}}
\caption{Results (\%) on the dusk-rainy scene. Here, the trained model is directly evaluated on the dusk-rainy scene.}\label{table7}
\vspace{-0.15in}
\end{table}

\begin{table}[ht]
\centering
\small
\scalebox{0.85}{
\tabcolsep=4.2pt
\begin{tabular}{l|ccccccc|c}
\toprule[1.5pt]
Method   & bus & bike & car  & motor  & person  & rider & truck & mAP  \\ \hline
Source Only   &23.4    &13.3    &31.8   &1.5     &10.2      &10.9     &23.2     &16.3 \\ \hline
DAF \cite{chen2018domain}     &24.2    &11.0    &32.4   &4.6     &12.7      &11.9     &27.7     &17.8 \\ \hline
CT \cite{zhao2020collaborative}      &19.5    &9.7     &29.0   &1.1     &9.9       &9.1      &17.6     &13.7 \\ \hline
SCL \cite{shen2019scl}     &22.9    &12.8    &35.8   &0.9     &14.8      &15.0     &30.2     &18.9 \\ \hline
SW \cite{saito2019strong}      &29.6    &10.4    &37.9   &0.7     &15.0      &11.1     &31.6     &19.5 \\ \hline
ICCR  \cite{xu2020exploring}   &28.4    &16.5    &33.6   &0.9     &16.4      &12.2     &30.3     &19.7 \\ \midrule
SW-VDD      &\textbf{35.7}    &\textbf{17.4}    &\textbf{42.2}   &\textbf{7.9}     &\textbf{18.1}      &\textbf{16.0}     &\textbf{33.9}     &\textbf{24.5} \\
\bottomrule[1.5pt]
\end{tabular}}
\caption{Results (\%) on the night-rainy scene. Here, the trained model is directly evaluated on the night-rainy scene.}\label{table8}
\vspace{-0.15in}
\end{table}

\subsection{Result Analysis of Compound-target DAOD}

For compound-target DAOD, we use the same optimization method as that of the single-target case. ResNet101 is the backbone. Table \ref{table6}, \ref{table7}, and \ref{table8} show the compared results. Here, the model trained on the compound-target DAOD is separately evaluated on the compound target, dusk-rainy, and night-rainy domain. Compared with SW \cite{saito2019strong}, plugging our disentanglement into SW improves its performance by 5.3\%, 5.6\%, and 5.0\%. Meanwhile, we can see that the performance of each category outperforms all compared methods significantly. This shows for single- and compound-target DAOD, extracting DIR is an efficient way. Meanwhile, the performance gain further demonstrates our method is capable of extracting DIR effectively.

\section{Conclusion}

In this paper, we propose vector-decomposed disentanglement for DAOD. We only defines an extractor to extract domain-invariant representations. Meanwhile, we do not use reconstruction to ensure the disentangled components contain all the information in the input. In the experiment, our method is separately evaluated on the single- and compound-target case. The performance gain over baselines shows the effectiveness of our method.

\section*{Acknowledgement}

This work is supported by the NSFC (under Grant 61876130, 61932009).

{\small
\bibliographystyle{ieee_fullname}
\bibliography{egbib}

\begin{thebibliography}{10}\itemsep=-1pt

\bibitem{bengio2013representation}
Yoshua Bengio, Aaron Courville, and Pascal Vincent.
\newblock Representation learning: A review and new perspectives.
\newblock {\em IEEE transactions on pattern analysis and machine intelligence},
  35(8):1798--1828, 2013.

\bibitem{ijcai2019-285}
Ruichu Cai, Zijian Li, Pengfei Wei, Jie Qiao, Kun Zhang, and Zhifeng Hao.
\newblock Learning disentangled semantic representation for domain adaptation.
\newblock In {\em Proceedings of the Twenty-Eighth International Joint
  Conference on Artificial Intelligence}, pages 2060--2066, 2019.

\bibitem{chen2020harmonizing}
Chaoqi Chen, Zebiao Zheng, Xinghao Ding, Yue Huang, and Qi Dou.
\newblock Harmonizing transferability and discriminability for adapting object
  detectors.
\newblock In {\em Proceedings of the IEEE/CVF Conference on Computer Vision and
  Pattern Recognition}, pages 8869--8878, 2020.

\bibitem{chen2018domain}
Yuhua Chen, Wen Li, Christos Sakaridis, Dengxin Dai, and Luc Van~Gool.
\newblock Domain adaptive faster r-cnn for object detection in the wild.
\newblock In {\em Proceedings of the IEEE conference on computer vision and
  pattern recognition}, pages 3339--3348, 2018.

\bibitem{cordts2016cityscapes}
Marius Cordts, Mohamed Omran, Sebastian Ramos, Timo Rehfeld, Markus Enzweiler,
  Rodrigo Benenson, Uwe Franke, Stefan Roth, and Bernt Schiele.
\newblock The cityscapes dataset for semantic urban scene understanding.
\newblock In {\em Proceedings of the IEEE conference on computer vision and
  pattern recognition}, pages 3213--3223, 2016.

\bibitem{do2019theory}
Kien Do and Truyen Tran.
\newblock Theory and evaluation metrics for learning disentangled
  representations.
\newblock {\em arXiv preprint arXiv:1908.09961}, 2019.

\bibitem{everingham2010pascal}
Mark Everingham, Luc Van~Gool, Christopher~KI Williams, John Winn, and Andrew
  Zisserman.
\newblock The pascal visual object classes (voc) challenge.
\newblock {\em International journal of computer vision}, 88(2):303--338, 2010.

\bibitem{ganin2014unsupervised}
Yaroslav Ganin and Victor Lempitsky.
\newblock Unsupervised domain adaptation by backpropagation.
\newblock {\em arXiv preprint arXiv:1409.7495}, 2014.

\bibitem{he2017mask}
Kaiming He, Georgia Gkioxari, Piotr Doll{\'a}r, and Ross Girshick.
\newblock Mask r-cnn.
\newblock In {\em Proceedings of the IEEE international conference on computer
  vision}, pages 2961--2969, 2017.

\bibitem{ResNet}
Kaiming He, Xiangyu Zhang, Shaoqing Ren, and Jian Sun.
\newblock Deep residual learning for image recognition.
\newblock In {\em Proceedings of the IEEE conference on computer vision and
  pattern recognition}, pages 770--778, 2016.

\bibitem{he2019multi}
Zhenwei He and Lei Zhang.
\newblock Multi-adversarial faster-rcnn for unrestricted object detection.
\newblock {\em arXiv preprint arXiv:1907.10343}, 2019.

\bibitem{he2020domain}
Zhenwei He and Lei Zhang.
\newblock Domain adaptive object detection via asymmetric tri-way faster-rcnn.
\newblock {\em European Conference on Computer Vision}, 2020.

\bibitem{higgins2018towards}
Irina Higgins, David Amos, David Pfau, Sebastien Racaniere, Loic Matthey,
  Danilo Rezende, and Alexander Lerchner.
\newblock Towards a definition of disentangled representations.
\newblock {\em arXiv preprint arXiv:1812.02230}, 2018.

\bibitem{inoue2018cross}
Naoto Inoue, Ryosuke Furuta, Toshihiko Yamasaki, and Kiyoharu Aizawa.
\newblock Cross-domain weakly-supervised object detection through progressive
  domain adaptation.
\newblock In {\em Proceedings of the IEEE conference on computer vision and
  pattern recognition}, pages 5001--5009, 2018.

\bibitem{kim2019self}
Seunghyeon Kim, Jaehoon Choi, Taekyung Kim, and Changick Kim.
\newblock Self-training and adversarial background regularization for
  unsupervised domain adaptive one-stage object detection.
\newblock In {\em Proceedings of the IEEE/CVF International Conference on
  Computer Vision}, pages 6092--6101, 2019.

\bibitem{kim2019diversify}
Taekyung Kim, Minki Jeong, Seunghyeon Kim, Seokeon Choi, and Changick Kim.
\newblock Diversify and match: A domain adaptive representation learning
  paradigm for object detection.
\newblock In {\em Proceedings of the IEEE Conference on Computer Vision and
  Pattern Recognition}, pages 12456--12465, 2019.

\bibitem{lee2018diverse}
Hsin-Ying Lee, Hung-Yu Tseng, Jia-Bin Huang, Maneesh Singh, and Ming-Hsuan
  Yang.
\newblock Diverse image-to-image translation via disentangled representations.
\newblock In {\em Proceedings of the European Conference on Computer Vision
  (ECCV)}, pages 35--51, 2018.

\bibitem{Li}
Shuang Li, Chi~Harold Liu, Xie Binhui, Limin Su, Zhengming Ding, and Gao Huang.
\newblock Joint adversarial domain adaptation.
\newblock In {\em Proceedings of the 27th ACM International Conference on
  Multimedia}, pages 729--737, 2019.

\bibitem{MM5}
Xin Li, Fan Yang, Hong Cheng, Junyu Chen, Yuxiao Guo, and Leiting Chen.
\newblock Multi-scale cascade network for salient object detection.
\newblock In {\em Proceedings of the 25th ACM International Conference on
  Multimedia}, page 439–447, 2017.

\bibitem{liu2020domain}
Feng Liu, Xiaoxong Zhang, Fang Wan, Xiangyang Ji, and Qixiang Ye.
\newblock Domain contrast for domain adaptive object detection.
\newblock {\em arXiv preprint arXiv:2006.14863}, 2020.

\bibitem{liu2016ssd}
Wei Liu, Dragomir Anguelov, Dumitru Erhan, Christian Szegedy, Scott Reed,
  Cheng-Yang Fu, and Alexander~C Berg.
\newblock Ssd: Single shot multibox detector.
\newblock In {\em European conference on computer vision}, pages 21--37.
  Springer, 2016.

\bibitem{liu2018detach}
Yen-Cheng Liu, Yu-Ying Yeh, Tzu-Chien Fu, Sheng-De Wang, Wei-Chen Chiu, and
  Yu-Chiang Frank~Wang.
\newblock Detach and adapt: Learning cross-domain disentangled deep
  representation.
\newblock In {\em Proceedings of the IEEE Conference on Computer Vision and
  Pattern Recognition}, pages 8867--8876, 2018.

\bibitem{liu2020open}
Ziwei Liu, Zhongqi Miao, Xingang Pan, Xiaohang Zhan, Dahua Lin, Stella~X Yu,
  and Boqing Gong.
\newblock Open compound domain adaptation.
\newblock In {\em Proceedings of the IEEE/CVF Conference on Computer Vision and
  Pattern Recognition}, pages 12406--12415, 2020.

\bibitem{locatello2018challenging}
Francesco Locatello, Stefan Bauer, Mario Lucic, Sylvain Gelly, Bernhard
  Sch{\"o}lkopf, and Olivier Bachem.
\newblock Challenging common assumptions in the unsupervised learning of
  disentangled representations.
\newblock 2019.

\bibitem{peng2019domain}
Xingchao Peng, Zijun Huang, Ximeng Sun, and Kate Saenko.
\newblock Domain agnostic learning with disentangled representations.
\newblock {\em ICML}, 2019.

\bibitem{redmon2016you}
Joseph Redmon, Santosh Divvala, Ross Girshick, and Ali Farhadi.
\newblock You only look once: Unified, real-time object detection.
\newblock In {\em Proceedings of the IEEE conference on computer vision and
  pattern recognition}, pages 779--788, 2016.

\bibitem{ren2015faster}
Shaoqing Ren, Kaiming He, Ross Girshick, and Jian Sun.
\newblock Faster r-cnn: Towards real-time object detection with region proposal
  networks.
\newblock In {\em Advances in neural information processing systems}, pages
  91--99, 2015.

\bibitem{ridgeway2018learning}
Karl Ridgeway and Michael~C Mozer.
\newblock Learning deep disentangled embeddings with the f-statistic loss.
\newblock In {\em Advances in Neural Information Processing Systems}, pages
  185--194, 2018.

\bibitem{saito2019strong}
Kuniaki Saito, Yoshitaka Ushiku, Tatsuya Harada, and Kate Saenko.
\newblock Strong-weak distribution alignment for adaptive object detection.
\newblock In {\em Proceedings of the IEEE Conference on Computer Vision and
  Pattern Recognition}, pages 6956--6965, 2019.

\bibitem{sakaridis2018semantic}
Christos Sakaridis, Dengxin Dai, and Luc Van~Gool.
\newblock Semantic foggy scene understanding with synthetic data.
\newblock {\em International Journal of Computer Vision}, 126(9):973--992,
  2018.

\bibitem{shao2018feature}
Rui Shao, Xiangyuan Lan, and Pong~C Yuen.
\newblock Feature constrained by pixel: Hierarchical adversarial deep domain
  adaptation.
\newblock In {\em Proceedings of the 26th ACM international conference on
  Multimedia}, pages 220--228, 2018.

\bibitem{shen2019scl}
Zhiqiang Shen, Harsh Maheshwari, Weichen Yao, and Marios Savvides.
\newblock Scl: Towards accurate domain adaptive object detection via gradient
  detach based stacked complementary losses.
\newblock {\em arXiv preprint arXiv:1911.02559}, 2019.

\bibitem{VGG}
Karen Simonyan and Andrew Zisserman.
\newblock Very deep convolutional networks for large-scale image recognition.
\newblock {\em arXiv preprint arXiv:1409.1556}, 2014.

\bibitem{su2020adapting}
Peng Su, Kun Wang, Xingyu Zeng, Shixiang Tang, Dapeng Chen, Di Qiu, and
  Xiaogang Wang.
\newblock Adapting object detectors with conditional domain normalization.
\newblock {\em European Conference on Computer Vision}, 2020.

\bibitem{wang2018visual}
Jindong Wang, Wenjie Feng, Yiqiang Chen, Han Yu, Meiyu Huang, and Philip~S Yu.
\newblock Visual domain adaptation with manifold embedded distribution
  alignment.
\newblock In {\em Proceedings of the 26th ACM international conference on
  Multimedia}, pages 402--410, 2018.

\bibitem{wang2019few}
Tao Wang, Xiaopeng Zhang, Li Yuan, and Jiashi Feng.
\newblock Few-shot adaptive faster r-cnn.
\newblock In {\em Proceedings of the IEEE Conference on Computer Vision and
  Pattern Recognition}, pages 7173--7182, 2019.

\bibitem{wu2021instance}
Aming Wu, Yahong Han, Linchao Zhu, and Yi Yang.
\newblock Instance-invariant domain adaptive object detection via progressive
  disentanglement.
\newblock {\em IEEE Transactions on Pattern Analysis and Machine Intelligence},
  2021.
\newblock doi: {10.1109/TPAMI.2021.3060446}.

\bibitem{xie2019multi}
Rongchang Xie, Fei Yu, Jiachao Wang, Yizhou Wang, and Li Zhang.
\newblock Multi-level domain adaptive learning for cross-domain detection.
\newblock {\em arXiv preprint arXiv:1907.11484}, 2019.

\bibitem{xu2020exploring}
Chang-Dong Xu, Xing-Ran Zhao, Xin Jin, and Xiu-Shen Wei.
\newblock Exploring categorical regularization for domain adaptive object
  detection.
\newblock In {\em Proceedings of the IEEE/CVF Conference on Computer Vision and
  Pattern Recognition}, pages 11724--11733, 2020.

\bibitem{xu2020cross}
Minghao Xu, Hang Wang, Bingbing Ni, Qi Tian, and Wenjun Zhang.
\newblock Cross-domain detection via graph-induced prototype alignment.
\newblock In {\em Proceedings of the IEEE/CVF Conference on Computer Vision and
  Pattern Recognition}, pages 12355--12364, 2020.

\bibitem{yu2018bdd100k}
Fisher Yu, Wenqi Xian, Yingying Chen, Fangchen Liu, Mike Liao, Vashisht
  Madhavan, and Trevor Darrell.
\newblock Bdd100k: A diverse driving video database with scalable annotation
  tooling.
\newblock {\em arXiv preprint arXiv:1805.04687}, 2018.

\bibitem{MM3}
Jiahui Yu, Yuning Jiang, Zhangyang Wang, Zhimin Cao, and Thomas Huang.
\newblock Unitbox: An advanced object detection network.
\newblock In {\em Proceedings of the 24th ACM international conference on
  Multimedia}, pages 516--520, 2016.

\bibitem{zhao2020collaborative}
Ganlong Zhao, Guanbin Li, Ruijia Xu, and Liang Lin.
\newblock Collaborative training between region proposal localization and
  classification for domain adaptive object detection.
\newblock In {\em European Conference on Computer Vision}, pages 86--102.
  Springer, 2020.

\bibitem{zhao2020adaptive}
Zhen Zhao, Yuhong Guo, Haifeng Shen, and Jieping Ye.
\newblock Adaptive object detection with dual multi-label prediction.
\newblock In {\em European Conference on Computer Vision}, pages 54--69.
  Springer, 2020.

\bibitem{zhu2019adapting}
Xinge Zhu, Jiangmiao Pang, Ceyuan Yang, Jianping Shi, and Dahua Lin.
\newblock Adapting object detectors via selective cross-domain alignment.
\newblock In {\em Proceedings of the IEEE Conference on Computer Vision and
  Pattern Recognition}, pages 687--696, 2019.

\bibitem{zhuo2017deep}
Junbao Zhuo, Shuhui Wang, Weigang Zhang, and Qingming Huang.
\newblock Deep unsupervised convolutional domain adaptation.
\newblock In {\em Proceedings of the 25th ACM international conference on
  Multimedia}, pages 261--269, 2017.

\end{thebibliography}
}

\end{document}